\documentclass[11pt]{article}

\usepackage[preprint]{acl}

\usepackage{times}
\usepackage{latexsym}

\usepackage[T1]{fontenc}

\usepackage[utf8]{inputenc}

\usepackage{microtype}

\usepackage{inconsolata}

\usepackage{graphicx}
\usepackage{subcaption}

\usepackage{balance}
\usepackage{enumitem}
\usepackage{multirow}
\usepackage[table]{xcolor}

\usepackage{acronym}

\usepackage{threeparttable}

\newcommand{\argmax}{\mathop{\mathrm{argmax}}\limits}

\usepackage{algorithm}      
\usepackage{algpseudocode}  
\usepackage{algorithmicx}   
\usepackage{amsmath}
\usepackage{amssymb}  

\usepackage{rotating}
\usepackage{tabularx}
\usepackage{booktabs}
\usepackage{adjustbox}
\usepackage{multirow}
\usepackage{dsfont}

\title{EmoMAS: Emotion-Aware Multi-Agent System for High-Stakes Edge-Deployable Negotiation with Bayesian Orchestration}

\author{
 \textbf{Yunbo Long\textsuperscript{1}},
 \textbf{Yuhan Liu\textsuperscript{2}},
 \textbf{Liming Xu\textsuperscript{1}},
\\
\textsuperscript{1}University of Cambridge, UK,
 \textsuperscript{2}University of Toronto, Canada
\\
}

\begin{document}
\maketitle
\begin{abstract}
Large language models (LLMs) has been widely used for automated negotiation, but their high computational cost and privacy risks limit deployment in privacy-sensitive, on-device settings such as mobile assistants or rescue robots. 
Small language models (SLMs) offer a viable alternative, yet struggle with the complex emotional dynamics of high-stakes negotiation.
We introduces \textbf{EmoMAS}, a Bayesian multi-agent framework that transforms emotional decision-making from reactive to strategic. 
EmoMAS leverages a Bayesian orchestrator to coordinate three specialized agents: game-theoretic, reinforcement learning, and psychological coherence models. The system fuses their real-time insights to optimize emotional state transitions while continuously updating agent reliability based on negotiation feedback.
This mixture-of-agents architecture enables online strategy learning without pre-training.
We further introduce four high-stakes, edge-deployable negotiation benchmarks across debt, healthcare, emergency response, and educational domains.
Through extensive agent-to-agent simulations across all benchmarks, both SLMs and LLMs equipped with EmoMAS consistently surpass all baseline models in negotiation performance while  balancing ethical behavior.
These results show that strategic emotional intelligence is also the key driver of negotiation success.
By treating emotional expression as a strategic variable within a Bayesian multi-agent optimization framework, EmoMAS establishes a new paradigm for effective, private, and adaptive negotiation AI suitable for high-stakes edge deployment.
\end{abstract}

\begin{figure*}[t]
    \centering
    \includegraphics[width=\textwidth]{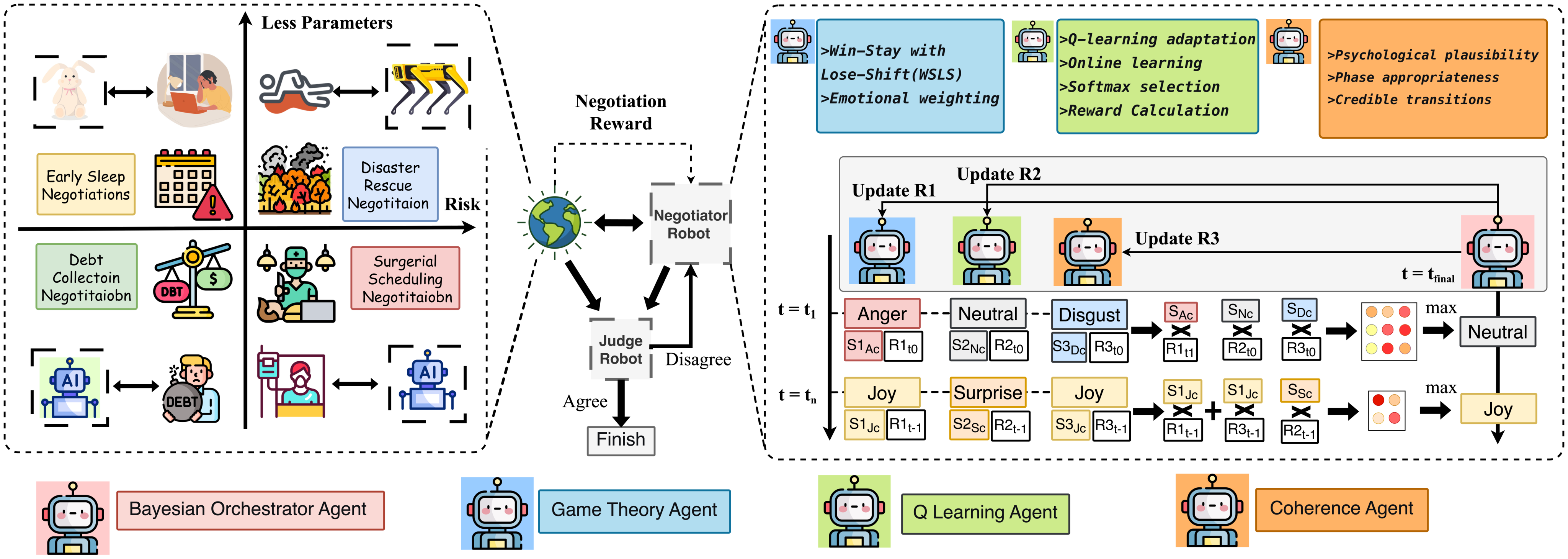}
    \caption{Illustration of the workflow of the EmoMAS framework.}
    \label{fig:emomas_framework}
\end{figure*}

\section{Introduction}

Large language models (LLMs) are increasingly deployed as negotiation agents \citep{zhu2025automated,long2025evoemoevolvedemotionalpolicies}, but their cloud-centric paradigm exposes sensitive negotiations to privacy and security risks. An edge-deployable alternative is urgently needed, where small language models (SLMs) can negotiate locally on devices and private infrastructure—protecting data in domains like mobile commerce, embodied robotics, and institutional bargaining. However, current SLM agents lack the strategic emotional intelligence for adversarial negotiations \citep{belcak2025small}. Trained on smaller emotional corpora, they fail to match the sophisticated emotional expression of LLMs, making them vulnerable to emotional manipulation \citep{hu2024llm,orpek2024language}. 
This weakness is magnified in the very scenarios that make edge deployment essential like high-stakes, emotionally-sensitive negotiations where data cannot leave local devices. In debt collection, a SLM agent must navigate genuine pleas while detecting emotional fraud. In hospital surgery scheduling, it must negotiate with patients and families under intense emotional duress to allocate limited resources. Educational companion robots need to defuse children's bedtime anxiety through compassionate dialogue, while disaster rescue robots must conduct psychological persuasion and negotiation with distressed victims in bandwidth-denied environments. These privacy-critical domains offer immense opportunity for SLM deployment, but they demand a level of strategic emotional intelligence that current small models fundamentally lack.

Existing LLM-based methods for emotional optimization face significant practical limitations. They typically require extensive pre-training, such as RL-based learning of emotional transitions across diverse scenarios, which is time-consuming and data-hungry. Recent advances in Mixture-of-Agents (MoA) and Mixture-of-Experts (MoE) frameworks offer a promising direction by coordinating multiple specialized models for enhanced reasoning \citep{wang2024mixture}. However, these approaches are similarly constrained by their reliance on static questions with targeted answers, which cannot adapt within a single, unfolding interaction. This approach is poorly suited to high-stakes edge scenarios where sensitive negotiation data is inherently scarce. Moreover, these learned strategies are often brittle and overfitted to the stable personality profiles of specific training agents, failing to generalize to new individuals with different emotional expressions. Consequently, when encountering a novel debtor or a new negotiation context, the system requires retraining from scratch. This challenge necessitates an online learning paradigm---an agent capable of optimizing emotional strategies in situ, without pre-training, by adapting to each unique opponent in real-time.

To address these gaps, we propose \textbf{EmoMAS}, a Bayesian multi-agent optimization framework for strategic emotional negotiation. Unlike static aggregation methods, our core innovation is a Bayesian Orchestrator that serves as a \textit{meta-reasoner}. 
It dynamically re-weights the predictions of three specialized agents---responsible for game-theoretic payoff, RL-based pattern learning, and psychological coherence---based on their context-specific reliability, which is learned online through Bayesian updating. 
This enables EmoMAS to optimize not just isolated emotional responses, but coherent emotional trajectories across uncertain-length dialogues, directly targeting the sparse, final reward of a successful agreement.
Through extensive agent-to-agent simulations across emotionally sensitive domains---credit finance (debt collection), healthcare (surgical scheduling), and disaster response (resource allocation)---we demonstrate that EmoMAS-equipped negotiators achieve superior performance in balancing strategic objectives with emotional coherence. 
Our framework shows that strategic emotional intelligence through Bayesian multi-agent optimization, rather than static emotional personas, is the critical factor for negotiation success in adversarial, emotionally charged environments.

\section{Related Work}\label{sec:related_work}

\subsection{Edge-Deployable Negotiation}

Autonomous LLM agents have been increasingly applied to role-playing scenarios such as card games, trading, and debt collection, where they simulate negotiating parties \citep{light2023avalonbench,longemodebt}. However, most existing work implicitly assumes cloud-based LLM agents with direct access sensitive information from banks, hospitals, or personal devices during negotiations, overlooking critical privacy and security risks associated with prompt-based information transmission strategies \citep{he2024emerged}. Moreover, these approaches rely heavily on persistent network connectivity and are susceptible to latency and reliability issues, which can significantly degrade user experience in time-sensitive \citep{belcak2025small}.
These limitations are especially acute in settings constrained by geopolitics or organisational policies, where access to LLM APIs may be restricted or unavailable. In remote areas with limited connectivity, or in embodied AI systems where robot swarms must negotiate with humans in real-time, reliance on cloud services becomes a fundamental bottleneck. These underscore the urgent need for offline, lightweight agents deployable on edge devices, enabling robust and private negotiations without external dependencies while ensuring data sovereignty and operational resilience across diverse  conditions.

\subsection{Small Language Models in Negotiation}

The paradigm of language models is broadly divided into LLMs and SLMs, with the latter typically defined as models with 7 billion parameters or fewer \citep{belcak2025small}. While LLMs have demonstrated remarkable, emergent abilities in general tasks, their massive scale induces critical limitations, including high computational demands, privacy concerns from cloud dependency, and unsuitability for real-time, edge-device applications \citep{orpek2024language}. Consequently, SLMs have gained prominence for their low latency, cost-effectiveness, and ease of customization. However, a well-documented performance gap persists, primarily attributed to the scaling laws; SLMs inherently lack the extensive world knowledge and nuanced reasoning capabilities of LLMs \citep{lu2024small,long2025eq}. 
While the limitations of SLMs in mathematical and commonsense reasoning are well established, their capacity for emotional intelligence---particularly in adopting emotional personas, inferring others’ emotions, and adapting strategies in real time during socio-emotional interactions such as negotiation---remains largely \textit{unexplored}.

\subsection{Multi-Agent Systems for Negotiation}

MoA and MoE architectures have proven effective for enhancing complex reasoning, where multiple specialized agents (or ``experts'') are leveraged to improve answer quality on tasks like mathematics or open-ended generation \citep{yan2025mathagent}. A core assumption in these frameworks is that a static aggregation of agent outputs
---
often via fixed averaging or voting
---
is sufficient, as the confidence and reliability of each agent's reasoning are treated as constant. Besides,  numerous methods are used to learn optimized weightings of agents in the  multi-agents systems for fixed problems, these are inherently pre-trained under the same questions and target answers.
However, these approaches face fundamental limitations when applied to long-horizon, emotionally dynamic negotiations. First, negotiation is a sequential decision-making process under uncertainty, where the optimal contribution of each strategic perspective (e.g., game theory vs. psychological coherence) must shift in real-time as the dialogue unfolds. Second, the reward structure is sparse and delayed; success is determined only at the conclusion of a variable-length interaction, not at each turn. This makes it impossible to optimize each step independently, requiring a framework that plans emotional trajectories toward the final outcome. Third, the reliability of each expert is context-dependent
---
an agent skilled at detecting deception may be crucial when facing a manipulative opponent but less so during rapport building. 
Existing MoA/MoE architectures lack the mechanism to learn and adapt these reliability weights online within a single negotiation.

\section{The EmoMAS Framework}

EmoMAS (\autoref{fig:emomas_framework}) is a Bayesian multi-agent system that strategically optimizes emotional transitions in negotiations by implementing a MoA architecture with online learning. 
Three specialized agents---a Game Theory agent for payoff optimization, a Reinforcement Learning (RL) agent for adaptive strategy learning, and a Coherence agent for psychological consistency---provide probabilistic predictions about optimal emotional state transitions. A Bayesian Orchestrator Agent serves as a meta-reasoner, dynamically weighting these predictions using context-specific reliability estimates that are updated in real-time via Bayesian inference, enabling the system to learn optimal emotional strategies during each negotiation without pre-training. This framework optimizes not just individual emotional responses but coherent emotional trajectories toward the sparse final reward of successful agreement, adapting agent contributions to the unfolding dialogue context through probabilistic fusion.

\subsection{In-Context Emotion Recognition}

EmoMAS performs emotion recognition through in-context learning, eliminating the need for task-specific fine-tuning.
For each debtor utterance, the system constructs a structured prompts comprising: 
(1) definitions of seven emotional states, 
(2) conversational examples, and 
(3) current dialogue context (see details in \autoref{app:prompts}). 
EmoMAS tracks debtor and creditor emotional trajectories using $\mathcal{H}_t^d = (D_{t-n}, \dots, D_t)$ and $\mathcal{H}_t^c = (C_{t-n}, \dots, C_t)$, respectively, with emotional states draw from $\mathcal{E}=\{\text{joy}, \text{sadness}, \text{anger}, \text{fear}, \text{disgust}, \text{surprise}, \text{neutral}\}$.

\subsection{Game Theory Agent}
This agent implements a Win-Stay, Lose-Shift (WSLS) strategy with emotional weighting. 
Instead of using pure Tit-for-Tat, our WSLS strategy (\autoref{tab:complete_payoff}) maintains cooperation for positive debtor emotions (joy, neutral, surprise) and shifts to cautious responses for negative exchanges (anger, disgust, fear). This avoids escalation risks while providing necessary resistance. 
This agent computes:
\begin{equation}
f_{\text{Payoff}}(d) = \argmax_{e \in \mathcal{E}} \pi(d, e)_2
\end{equation}
with $\pi(d, e) = (\pi_1(d,e), \pi_2(d,e))$ representing payoffs between the negotiators and opponents, respectively (\autoref{app:wsls}).

\subsection{Reinforcement Learning Agent}
We adopt a Q-learning approach to enable online adaptation without the need for neural network training.
The agent maintains a Q-table $Q(s, a)$, where each state $s$ corresponds to a discretized representation of the negotiation context. 
Specifically, at time $t$, the state is defined as $s_t = \langle e^c_t, e^d_t, \phi_t, g_t \rangle$, where $e^c_t$ denotes the emotion of the creditor, $e^d_t$ the emotion of the debtor, $\phi_t$ the negotiation phase (early, middle, late, or crisis), and $g_t$ a categorical indicator of the gap size (small or large). 
Actions correspond to emotional responses $a_t \in \mathcal{E}$. 
Updates follow the Q-learning rule:
\begin{equation}
    \begin{split}
    Q(s_t, a_t) \leftarrow Q(s_t, a_t) + \\ \alpha [ R_{t+1} + \gamma \max_{a'} Q(s_{t+1}, a') - Q(s_t, a_t) ]
    \end{split}
\end{equation}
where reward $R_t$ combines negotiation success and efficiency, $\alpha=0.1$ is learning rate, and $\gamma=0.9$ is the discount factor. 
Emotion selection uses softmax with temperature $\tau=0.1$:
\begin{equation}
\pi(a|s) = \frac{\exp(Q(s,a)/\tau)}{\sum_{a'} \exp(Q(s,a')/\tau)}
\end{equation}
This tabular approach enables true online learning during individual negotiations
---
unlike DQN which requires offline training epochs
---making it suitable for rapid adaptation to specific debtor characteristics.
See details in \autoref{app:rl_method_analysis}.

\subsection{Emotional Coherence Agent}
The Emotional Coherence Agent employs LLM-based psychological reasoning to evaluate emotion transitions. Given a context vector \(\mathbf{c} = (e_c, e_d, \phi, r, g, d, \mathbf{h})\) representing current creditor emotion \(e_c\), debtor emotion \(e_d\), negotiation phase \(\phi\), round number \(r\), gap size \(g\), debt amount \(d\), and emotional history \(\mathbf{h}\), the \ac{LLM} outputs an assessment matrix \(\mathbf{A} \in \mathbb{R}^{7 \times 4}\) where each row corresponds to an emotion \(e \in \mathcal{E} = \{\text{joy}, \text{sadness}, \text{anger}, \text{fear}, \text{surprise}, \text{disgust}, \text{Neutral}\}\) with columns for plausibility \(p \in [0,1]\), appropriateness \(a \in [0,1]\), strategic value \(s \in [0,1]\), and psychological rationale \(r\). The agent computes selection probabilities via softmax normalization: 
\[
P(e_i) = \frac{\exp(f(p_i, a_i, s_i)/\tau)}{\sum_{j=1}^{7} \exp(f(p_j, a_j, s_j)/\tau)},
\]
where \(f(\cdot)\) aggregates dimension scores through LLM-guided weighting and \(\tau = 1.0\) controls exploration temperature. This formulation enables psychologically-grounded emotional transitions without hard-coded rules.
See details in \autoref{app:coherence-details}.

\begin{algorithm}[t]
\centering\small
\caption{EmoMAS Framework.}
\label{alg:EmoMAS}
\begin{algorithmic}[1]
\Procedure{Negotiate}{$\mathcal{M}_{o}, \mathcal{M}_{n}, \mathcal{M}_{j}, S$}
    \State Initialize $H \gets \emptyset$, $\mathcal{C} \gets \text{neutral}$, $e_n \gets \text{neutral}$, $e_d \gets S.\text{emotion}$
    \For{$t = 0$ to $T_{\max}$}
        \State $m_d \gets \mathcal{M}_d(H, e_d)$
        \State $e_d \gets \text{RecognizeEmotion}(m_d)$
        \State $H \gets H \cup \{(m_d, e_d)\}$
        
        \State $e_n \gets \text{SelectEmotion}(e_d, \mathcal{C}, H)$
        \State $m_n \gets \mathcal{M}_n(H, e_n)$
        \State $H \gets H \cup \{(m_n, e_n)\}$
        
        \If{$\mathcal{M}_j.\text{AgreementReached}(H)$} 
            \State \textbf{return} $\text{success}, H$
        \ElsIf{$\mathcal{M}_j.\text{NegotiationFailed}(H)$}
            \State \textbf{return} $\text{failure}, H$
        \EndIf
        \State $\mathcal{C} \gets \text{UpdateContext}(H, e_n)$
    \EndFor
    \State \textbf{return} $\text{timeout}, H$
\EndProcedure

\Function{SelectEmotion}{$e_d, \mathcal{C}, H$}
    \State $p_{\text{GT}} \gets \text{GameTheoryAgent}(e_d, \mathcal{C})$
    \State $p_{\text{RL}} \gets \text{RLAgent}(e_d, \mathcal{C}, H)$
    \State $p_{\text{EC}} \gets \text{CoherenceAgent}(e_d, \mathcal{C}, H)$
    \State \textbf{return} $\text{BayesianOrchestrator}(p_{\text{GT}}, p_{\text{RL}}, p_{\text{EC}})$
\EndFunction
\end{algorithmic}
\end{algorithm}

\subsection{Bayesian Orchestrator Agent}
The orchestrator integrates three expert agents' predictions through real-time Bayesian learning. It maintains reliability distributions \(\mathbf{w}^{(i)}_t\) for each agent \(i \in \{\text{GT}, \text{RL}, \text{Coherence}\}\), which evolve via Bayesian updating:
\begin{equation}
w^{(i)}_t \propto \underbrace{w^{(i)}_{t-1}}_{\text{prior}} \cdot \underbrace{\mathcal{L}\left(\mathbf{p}^{(i)}_{t-1} \mid \text{success}_{t-1}, \text{context}_t\right)}_{\text{likelihood}}
\end{equation}
where the likelihood function \(\mathcal{L}\) measures alignment between agent predictions and negotiation outcomes. Two reliability tracking mechanisms operate:

\paragraph{Macro-Level Reliability.} Updated after complete negotiation trajectories based on collection efficiency. For successful negotiations with collection rate \(\rho\), reliability increases by \(\Delta w^{(i)} \propto \rho \times \text{agent\_accuracy}^{(i)}\).

\paragraph{Micro-Level Reliability.} Within negotiations, weights adjust based on real-time prediction agreement with selected emotions.
The final emotion selection follows a weighted sum of reliability and confidence:
\begin{equation}
\text{Score}(e_j) = \sum_{i=1}^{3} w^{(i)}_t \cdot \text{confidence}^{(i)}(e_j)
\end{equation}
where \(\text{confidence}^{(i)}(e_j)\) is agent \(i\)'s confidence in emotion \(e_j\). The orchestrator strictly selects from the union of agents' recommendations:
\begin{equation}
e_{\text{selected}} = \argmax_{e_j \in \bigcup_i \mathcal{E}^{(i)}} \text{Score}(e_j)
\end{equation}
where \(\mathcal{E}^{(i)}\) is the set of emotions recommended by agent \(i\). This constraint ensures interpretability and respects each agent's expertise domain. 
Exploration occurs only through individual agents' exploration mechanisms, not via orchestrator-level random exploration. 
As a baseline, we implement a context-reasoning orchestrator that relies on LLM-based contextual reasoning to select emotionally appropriate responses, without probabilistic integration of specialized agents.

\subsection{Multi-Agent Negotiation Simulations}
The complete EmoMAS framework operates through an automated multi-agent simulation system, as formalized in \autoref{alg:EmoMAS}. The simulation involves three specialized agents: a \textit{oppoent agent} ($\mathcal{M}_{o}$) that generates client responses, a \textit{negotiator agent} ($\mathcal{M}_{n}$) that employs emotional intelligence strategies like EmoMAS, and a \textit{judge agent} ($\mathcal{M}_{j}$) that evaluates negotiation outcomes. Each round consists of emotional state recognition, multi-agent decision integration, and response generation guided by the selected emotional strategy. The system enables both real-time learning within negotiations and cumulative improvement across multiple scenarios.

\section{Experimental Settings}
\label{app:settings}

\begin{table*}[t]
\centering
\setlength{\cmidrulewidth}{0.2pt}  
\setlength{\aboverulesep}{1.5pt}
\setlength{\belowrulesep}{1.5pt}
\renewcommand{\arraystretch}{1}
\caption{
Performance comparison of EmoMAS and baselines (using GPT‑4o‑mini and Qwen‑7B) against vanilla GPT‑4o‑mini opponents across four scenarios (mean with 95\% confidence interval). Best results for each scenario are highlighted in \textbf{bold}.}
\label{tab:comprehensive_results}
\scalebox{0.75}{%
\begin{tabular}{p{1.6cm}lcp{1.5cm}lllll}
\toprule
\multirow{2}{*}{\textbf{Datasets}} & \multirow{2}{2cm}{\textbf{Negotiator Models}} & \multicolumn{2}{c}{\textbf{Success Rate (\%)} $\uparrow$} & \multicolumn{2}{c}{\textbf{Negotiation Outcomes} (\%) $\uparrow$} & \multicolumn{2}{c}{\textbf{Negotiation Rounds ($\downarrow$)}} \\
\cmidrule(lr){3-4} \cmidrule(lr){5-6} \cmidrule(lr){7-8}
 & & GPT-4o-mini & Qwen-7B & GPT-4o-mini & Qwen-7B & GPT-4o-mini & Qwen-7B \\
\midrule
\multirow{7}{1.75cm}{CRAD (Debt)} 
& Vanilla & 
90.0 &85.0  & 
14.5 \textcolor{gray}{[9.7-20.1]} & 12.6 \textcolor{gray}{[7.9-18.1]} & 
\bf{15.0} \textcolor{gray}{[13.1-17.7]} &17.6 \textcolor{gray}{[14.5-20.6]} \\

& Vanilla+Prompt & 
90.0  & 80.0 & 
14.5 \textcolor{gray}{[9.8-20.8]} & \bf{12.8} \textcolor{gray}{[8.1-18.5]} & 
15.7 \textcolor{gray}{[13.5-17.6]} & 17.4 \textcolor{gray}{[15.1-21.2]} \\

& Game Theory & 
95.0  & 70.0 & 
14.7 \textcolor{gray}{[9.9-20.5]} & 8.9  \textcolor{gray}{[4.7-13.9]} & 
15.1 \textcolor{gray}{[12.8-17.1]} & 20.1 \textcolor{gray}{[16.4-23.5]} \\

& Q-Learning & 
90.0 & 75.0  & 
14.2 \textcolor{gray}{[9.8-19.3]} &11.5 \textcolor{gray}{[6.5-17.5]} & 
15.8 \textcolor{gray}{[13.4-17.7]} & 16.5 \textcolor{gray}{[13.7-19.2]} \\

& Coherence & 
85.0 & 80.0  & 
13.8 \textcolor{gray}{[8.7-19.8]} & 12.4  \textcolor{gray}{[7.4-18.5]} & 
16.1 \textcolor{gray}{[14.9-21.5]} & 16.8 \textcolor{gray}{[13.9-19.5]} \\

\cmidrule(lr){2-8}
& \textbf{EmoMAS-LLM} & 
\bf{100.0} & 85.0  & 
14.3 \textcolor{gray}{[9.7-19.8]} & \bf{12.8} \textcolor{gray}{[8.2-18.6]} & 
15.3 \textcolor{gray}{[12.8-17.3]} & 16.6 \textcolor{gray}{[13.5-20.4]} \\

& \textbf{EmoMAS-Bayes} & 
\bf{100.0} & \bf{90.0}  & 
\bf{14.8} \textcolor{gray}{[10.0-21.2]} & 12.6 \textcolor{gray}{[8.1-18.4]} & 
15.2\textcolor{gray}{[14.4-16.0]} & \bf{16.4} \textcolor{gray}{[13.3-19.5]} \\

\midrule
\multirow{7}{1.75cm}{SSD (Medical)} 
& Vanilla & 
\bf{86.0} &68.0  & 
85.7 \textcolor{gray}{[57.1-100.0]} & 28.5 \textcolor{gray}{[0.0-57.1]} & 
\bf{7.7} \textcolor{gray}{[2.5-16.0]} &18.6 \textcolor{gray}{[9.2-27.3]} \\

& Vanilla+Prompt & 
77.0  & 69.0 & 
57.1 \textcolor{gray}{[14.3-85.7]} & 29.5 \textcolor{gray}{[0.1-57.3]} & 
11.1 \textcolor{gray}{[2.8-19.4]} & 21.2 \textcolor{gray}{[11.8-27.8]} \\

& Game Theory & 
45.0  & \bf{75.0} & 
42.8 \textcolor{gray}{[14.2-85.7]} & 2.2 \textcolor{gray}{[1.9-2.4]} & 
16.2 \textcolor{gray}{[6.1-26.8]} & \bf{7.9} \textcolor{gray}{[1.9-8.9]} \\

& Q-Learning & 
40.0 & 60.0  & 
8.2 \textcolor{gray}{[1.2-10.8]} & 3.7 \textcolor{gray}{[1.5-7.5]} & 
15.2 \textcolor{gray}{[8.9-20.1]} & 8.9 \textcolor{gray}{[0.5-14.4]} \\

& Coherence & 
46.0 & 54.0  & 
38.2 \textcolor{gray}{[10.1-79.6]} & 17.6 \textcolor{gray}{[0.6-50.4]} & 
12.8 \textcolor{gray}{[3.9-22.7]} & 20.5 \textcolor{gray}{[10.5-27.1]} \\

\cmidrule(lr){2-8}
& \textbf{EmoMAS-LLM} & 
80.0 & 60.0  & 
83.5 \textcolor{gray}{[56.2-100.0]} & 28.9 \textcolor{gray}{[0.5-57.3]} & 
10.2 \textcolor{gray}{[2.8-21.3]} & 17.1 \textcolor{gray}{[8.1-25.4]} \\

& \textbf{EmoMAS-Bayes} & 
84.0 & \bf{75.0}  & 
\bf{86.4} \textcolor{gray}{[59.2-100.0]} & \bf{33.7} \textcolor{gray}{[2.5-65.5]} & 
16.9 \textcolor{gray}{[6.8-27.4]} & 22.9 \textcolor{gray}{[12.6-29.4]} \\

\midrule
\multirow{7}{1.75cm}{DESRD (Emergency)} 
& Vanilla & 
25.0 &45.0  & 
21.6 \textcolor{gray}{[4.5-40.2]} & 41.5 \textcolor{gray}{[21.9-61.9]} & 
24.5 \textcolor{gray}{[19.4-28.6]} &19.5 \textcolor{gray}{[14.7-24.2]} \\

& Vanilla+Prompt & 
26.0  & 42.0 & 
22.7 \textcolor{gray}{[6.9-41.8]} & 45.2 \textcolor{gray}{[25.1-65.5]} & 
24.6 \textcolor{gray}{[19.2-29.5]} & 20.2 \textcolor{gray}{[15.3-25.0]} \\

& Game Theory & 
\bf{65.0}  & 52.0 & 
3.1 \textcolor{gray}{[2.0-3.5]} & 49.3 \textcolor{gray}{[29.3-70.4]} & 
\bf{9.2} \textcolor{gray}{[2.9-10.1]} & 20.4 \textcolor{gray}{[14.6-25.1]} \\

& Q-Learning & 
40.0 & \bf{60.0}  & 
16.8 \textcolor{gray}{[2.8-36.2]} & 54.2 \textcolor{gray}{[31.2-86.1]} & 
15.2 \textcolor{gray}{[8.9-20.1]} & 18.9 \textcolor{gray}{[14.5-24.4]} \\

& Coherence & 
46.0 & 36.0  & 
10.5 \textcolor{gray}{[3.1-24.6]} & 42.6 \textcolor{gray}{[22.5-63.1]} & 
18.9 \textcolor{gray}{[16.4-27.3]} & 19.3 \textcolor{gray}{[16.8-24.8]} \\

\cmidrule(lr){2-8}
& \textbf{EmoMAS-LLM} & 
56.0 & 55.0  & 
20.2 \textcolor{gray}{[3.2-39.8]} & 50.2 \textcolor{gray}{[29.1-72.5]} & 
16.1 \textcolor{gray}{[9.3-21.2]} & \bf{15.8} \textcolor{gray}{[11.4-19.6]} \\

& \textbf{EmoMAS-Bayes} & 
\bf{65.0} & \bf{60.0}  & 
\bf{26.7} \textcolor{gray}{[8.4-51.8]} & \bf{56.3} \textcolor{gray}{[33.1-87.6]} & 
19.5 \textcolor{gray}{[17.9-28.1]} & 20.7 \textcolor{gray}{[15.8-26.5]} \\

\midrule
\multirow{7}{1.75cm}{SSAD (Education)} 
& Vanilla & 
60.0 &36.0  & 
57.3 \textcolor{gray}{[37.0-77.3]} & 28.7 \textcolor{gray}{[9.5-48.3]} & 
14.4 \textcolor{gray}{[8.8-20.2]} &16.0 \textcolor{gray}{[10.2-21.9]} \\

& Vanilla+Prompt & 
\bf{75.0}  & 45.0 & 
71.3 \textcolor{gray}{[52.2-87.1]} & 41.9 \textcolor{gray}{[22.9-61.3]} & 
\bf{10.5} \textcolor{gray}{[6.1-15.5]} & 15.8 \textcolor{gray}{[9.9-21.8]} \\

& Game Theory & 
55.0  & 40.0 & 
52.4 \textcolor{gray}{[32.6-72.2]} & 37.6 \textcolor{gray}{[18.4-57.1]} & 
12.0 \textcolor{gray}{[6.7-17.6]} & \bf{15.5} \textcolor{gray}{[9.7-21.3]} \\

& Q-Learning & 
\bf{75.0} & 40.0  & 
70.9 \textcolor{gray}{[51.7-86.8]} & 36.9 \textcolor{gray}{[18.3-56.3]} & 
11.8 \textcolor{gray}{[7.0-17.2]} & 17.2 \textcolor{gray}{[11.2-23.1]} \\

& Coherence & 
46.0 & 30.0  & 
44.3 \textcolor{gray}{[26.2-67.2]} & 14.6 \textcolor{gray}{[4.2-36.5]} & 
17.2 \textcolor{gray}{[9.8-23.1]} & 18.9 \textcolor{gray}{[12.6-25.4]} \\

\cmidrule(lr){2-8}
& \textbf{EmoMAS-LLM} & 
80.0 & 50.0  & 
65.5 \textcolor{gray}{[42.2-85.8]} & \bf{42.3} \textcolor{gray}{[20.2-64.5]} & 
13.1 \textcolor{gray}{[8.2-18.3]} & 16.5 \textcolor{gray}{[10.6-22.8]} \\

& \textbf{EmoMAS-Bayes} & 
\bf{75.0} & \bf{60.0}  & 
\bf{75.6} \textcolor{gray}{[56.8-91.2]} & 40.5 \textcolor{gray}{[21.4-60.7]} & 
15.2 \textcolor{gray}{[8.9-20.1]} & 15.9 \textcolor{gray}{[10.2-21.8]} \\

\bottomrule
\end{tabular}%
}
\end{table*}

\paragraph{Datasets.}

We conduct experiments across four high-stakes, emotionally charged negotiation domains to evaluate generalization under conflicting needs. The primary domain uses the Credit Recovery Assessment Dataset (CRAD) \citep{long2025emodebtbayesianoptimizedemotionalintelligence} for debt negotiation. We further introduce three new benchmarks: (1) the Surgical Scheduling Dataset (SSD), which focuses on urgent medical negotiation involving surgical timing and constraints related to surgeon expertise; (2) the Disaster Emotional Support \& Rescue Dataset (DESRD), designed for emergency negotiation with injured survivors regarding rescue waiting times; and (3) the Student Sleep Alerting Dataset (SSAD), which addresses educational negotiation over bedtime under deadline pressure arising from academic or work-related commitments.
See details in \autoref{app:datasets}.

\paragraph{Agent Models.}
Considering negotiators may be deployed on robots, mobile devices, and institutional systems, we evaluate our approach across both small (SLMs) and large language models (LLMs) to assess its scalability and generalization.Specifically, the SLMs include Qwen-7B and Qwen-1.5B, representing different parameter scales within the open-weight Qwen family, while the LLMs include GPT-4o-mini as a representative commercial, closed-source model.

\paragraph{Baseline Models and Opponent Strategies.}
We compare our approach against five baseline systems representing distinct paradigms in negotiation: (1) Vanilla Single-Agent without emotional guidance, (2) Vanilla with Emotion Selection (prompt-guided emotional strategies), (3) Game Theory Agent (equilibrium-based reasoning), (4) RL Online Learning Agent (reward-driven adaptation),(5) Coherence Agent (psychological plausibility), and (6) Mixture of Agents Systems including EmoMAS-LLM (orchestrated by an LLM controller) and EmoMAS-Bayes (orchestrated by Bayesian inference). These baselines are applied to our primary agents (creditor, surgical scheduler, rescue robot, home robot), while their negotiation counterparts employ different strategies: a Vanilla Emotional baseline alongside three psychologically-informed advanced strategies—Pressure Tactics (deadlines, scarcity cues), Victim Playing (appeals to sympathy, learned helplessness), and Threatening Strategies (ultimatums, consequence escalation)—creating a comprehensive testbed for evaluating robustness against diverse behaviors.

\paragraph{Experimental Design.}
Our evaluation consists of three systematic experiments: 
(1) Baseline Comparison: We compare all baseline systems against our EmoMAS methods across all four scenarios under the vanilla opponent strategy. 
(2) Robustness Test: We repeat the same comparison against three advanced opponent strategies (pressure tactics, victim playing, and threatening strategies) on both medical and educational scenarios to assess robustness of different methods under adversarial conditions. 
(3) Model‑Scale Analysis: We compare those methods with edge-deployable SLMs and Cloud-based LLMs on emergency scenario, isolating the effect of strategic sophistication and model scale on negotiation outcomes. 
(4) Behavior Evaluations: We assess three ethical dimensions—manipulation behavior, emotional instruction following accuracy, and emotional consistency—on emergency scenario when facing advanced opponent strategies. Additionally, our results also serve as the ablation study, as most baselines represent key components integrated into our EmoMAS framework.
See all the prompts in the \autoref{app:prompts}.
And the values of hyperparameter are specified in \autoref{app:hyperparameters}.

\begin{table*}[t]
\centering\small
\setlength{\cmidrulewidth}{0.2pt}  
\setlength{\aboverulesep}{1.5pt}
\setlength{\belowrulesep}{1.5pt}
\renewcommand{\arraystretch}{1.0}
\setlength{\tabcolsep}{3pt} 
\caption{Performance comparison of EmoMAS and baselines (using GPT‑4o‑mini) against GPT‑4o‑mini opponents employing advanced strategies across medical and educational scenarios (mean with 95\% confidence interval). 
Best values for each scenario-opponent strategy combination are highlighted in \textbf{bold}.}
\label{tab:opponents_emotion}
\begin{tabular}{llcclllll}
\toprule
\multirow{2}{2cm}{\textbf{Opponent Strategies}} & \multirow{2}{2cm}{\textbf{Negotiator Models}} & \multicolumn{2}{c}{\textbf{Success Rate (\%)} $\uparrow$} & \multicolumn{2}{c}{\textbf{Negotiation Outcomes}(\%) $\uparrow$} & \multicolumn{2}{c}{\textbf{Negotiation Rounds} ($\downarrow$)} \\
\cmidrule(lr){3-4} \cmidrule(lr){5-6} \cmidrule(lr){7-8}
 & & SSD & SSAD & SSD & SSAD & SSD & SSAD \\
\midrule
\multirow{4}{*}{Pressuring} 
& Vanilla & 
20.0 &70.0  & 
18.8 \textcolor{gray}{[11.3-27.1]} & 1.9 \textcolor{gray}{[1.2-2.6]} & 
13.7 \textcolor{gray}{[10.8-16.6]} & \bf{7.6} \textcolor{gray}{[2.6-11.0]} \\

& Game Theory & 
32.0 &64.0  & 
14.1 \textcolor{gray}{[8.5-21.8]} & 0.6 \textcolor{gray}{[0.3-1.1]} & 
\bf{12.5} \textcolor{gray}{[9.8-15.2]} &8.9 \textcolor{gray}{[3.8-13.6]} \\

& Q-Learning & 
42.0 &76.0  & 
24.0 \textcolor{gray}{[14.5-35.1]} & 0.3 \textcolor{gray}{[0.1-0.8]} & 
15.6 \textcolor{gray}{[12.8-19.4]} &8.9 \textcolor{gray}{[4.1-15.6]} \\

& \textbf{EmoMAS-Bayes} & 
\bf{50.0} & \bf{80.0}  & 
\bf{28.0} \textcolor{gray}{[17.8-42.1]} & \bf{2.4} \textcolor{gray}{[1.7-3.8]} & 
13.8 \textcolor{gray}{[10.5-15.2]} &7.8 \textcolor{gray}{[1.8-8.6]} \\

\midrule
\multirow{4}{*}{Playing Victim} 
& Vanilla & 
58.0  & \bf{80.0} & 
55.4 \textcolor{gray}{[44.6-65.8]} & 1.7 \textcolor{gray}{[1.2-2.9]} & 
\bf{11.2} \textcolor{gray}{[8.5-14.0]} & 5.5 \textcolor{gray}{[2.6-8.8]} \\

& Game Theory & 
50.0  & 76.0 & 
47.6 \textcolor{gray}{[38.8-56.4]} & 0.8 \textcolor{gray}{[0.4-1.7]} & 
13.6 \textcolor{gray}{[10.5-16.3]} & 6.4 \textcolor{gray}{[3.3-10.1]} \\

& Q-Learning & 
28.0  & 68.0 & 
50.7 \textcolor{gray}{[39.6-60.8]} & 0.8 \textcolor{gray}{[0.3-1.6]} & 
14.2 \textcolor{gray}{[11.1-17.1]} & 7.5 \textcolor{gray}{[2.4-10.6]} \\

& \textbf{EmoMAS-Bayes} & 
\bf{70.0}  & \bf{80.0} & 
\bf{58.7} \textcolor{gray}{[46.1-68.3]} & \bf{2.1} \textcolor{gray}{[1.5-3.1]} & 
12.1 \textcolor{gray}{[9.6-14.5]} & \bf{4.4} \textcolor{gray}{[2.1-6.2]} \\

\midrule
\multirow{4}{*}{Threatening  } 
& Vanilla & 
70.0  & 76.0 & 
66.7 \textcolor{gray}{[56.7-76.2]} & 2.2 \textcolor{gray}{[1.9-2.4]} & 
11.8 \textcolor{gray}{[9.2-14.5]} & 7.9 \textcolor{gray}{[1.9-8.9]} \\

& Game Theory & 
42.0  & 70.0 & 
21.5 \textcolor{gray}{[13.1-31.4]} & 1.9 \textcolor{gray}{[1.6-2.3]} & 
10.8 \textcolor{gray}{[7.5-13.1]} & \bf{7.2} \textcolor{gray}{[1.6-8.1]} \\

& Q-Learning & 
64.0  & \bf{80.0} & 
68.7 \textcolor{gray}{[60.1-81.5]} & \bf{2.5} \textcolor{gray}{[2.2-3.3} & 
16.2 \textcolor{gray}{[13.1-20.2]} & 8.5 \textcolor{gray}{[2.4-9.9]} \\

& \textbf{EmoMAS-Bayes} & 
\bf{80.0}  & 75.0 & 
\bf{70.1} \textcolor{gray}{[66.3-83.6]} & 2.2 \textcolor{gray}{[1.9-2.4]} & 
\bf{10.2} \textcolor{gray}{[7.1-12.5]} & 8.3 \textcolor{gray}{[2.1-9.1]} \\

\bottomrule
\end{tabular}%
\end{table*}

\begin{table*}[h!]
\centering\small
\setlength{\cmidrulewidth}{0.2pt}  
\setlength{\aboverulesep}{1.5pt}
\setlength{\belowrulesep}{1.5pt}
\renewcommand{\arraystretch}{1.0}
\setlength{\tabcolsep}{3pt} 
\caption{Evaluation results of EmoMAS and baselines (Qwen-1.5B) against GPT‑4o‑mini and Qwen-1.5B opponents under the emergency scenario (mean with 95\% confidence interval). Best results for each opponent model are highlighted in \textbf{bold}.}
\label{tab:different_architecture}
\begin{tabular}{llclll}
\toprule
\multirow{2}{2cm}{\textbf{Negotiator Models}} & \multirow{2}{2cm}{\textbf{Opponent Strategies}} & \multicolumn{3}{c}{\textbf{Opponent Model (Qwen-1.5B)}} \\
\cmidrule(lr){3-5}
 & & \textbf{Success Rate (\%)} $\uparrow$ & \textbf{Negotiate Rates (\%)} $\uparrow$& \textbf{Negotiation Rounds} ($\downarrow$) \\
\midrule
\multirow{6}{*}{Qwen-1.5B}

& Vanilla & 90.0 & 91.5 \textcolor{gray}{[81.5-100.0]} & \bf{9.6} \textcolor{gray}{[6.1-13.2]} \\

& Coherence & 72.0 & 85.1 \textcolor{gray}{[57.4-100.0]} & 14.1 \textcolor{gray}{[10.1-19.9]} \\

& Game Theory & 86.0 & 88.7 \textcolor{gray}{[73.7-100.0]} & 11.9 \textcolor{gray}{[8.3-15.8]} \\

& Q-Learning & 94.0 & 90.9 \textcolor{gray}{[79.6-100.0]} &12.9 \textcolor{gray}{[9.3-16.1]} \\

& \textbf{EmoMAS-LLM} & 92.0 & 89.5 \textcolor{gray}{[69.8-100.0]} & 9.8 \textcolor{gray}{[6.6-15.4]} \\

& \textbf{EmoMAS-Bayes} & \bf{100.0} & \bf{99.5} \textcolor{gray}{[97.0-100.0]} & 10.2 \textcolor{gray}{[6.4-14.0]} \\

\midrule
\multirow{6}{*}{GPT-4o-mini}

& Vanilla & 98.0 & 96.7 \textcolor{gray}{[89.9-100.0]} & 11.9 \textcolor{gray}{[8.2-15.7]} \\

& Coherence & 78.0 & 86.2 \textcolor{gray}{[96.9-100.0]} & 13.7 \textcolor{gray}{[7.1-21.4]} \\

& Game Theory & 96.0 & 98.2 \textcolor{gray}{[96.9-100.0]} & 8.7 \textcolor{gray}{[5.1-13.1]} \\

&Q-Learning & 92.0 & 83.3 \textcolor{gray}{[63.3-100.0]} & 9.3 \textcolor{gray}{[4.5-15.3]} \\

& \textbf{EmoMAS-LLM} & \bf{100.0} & \bf{98.5} \textcolor{gray}{[97.9-100.0]} & 12.8 \textcolor{gray}{[8.2-14.9]} \\

& \textbf{EmoMAS-Bayes} & 98.0 & 96.9 \textcolor{gray}{[91.2-100.0]} & \bf{7.8} \textcolor{gray}{[3.8-11.5]} \\

\bottomrule
\end{tabular}%
\end{table*}

\paragraph{Evaluation Metrics.}
We evaluate negotiation performance using three core metrics: \textit{success rate} (proportion of successful agreements), \textit{negotiation outcomes} (cost/time reduction/increase relative to opponents target values), and \textit{negotiation rounds} (dialogue turns until resolution). 
For each scenario type---debt, medical, emergency, and education---we report mean values with 95\% confidence intervals computed via t-distribution, with non-negative bounds enforced for inherently positive metrics. 
See details in \autoref{app:datasets}.
All results are aggregated over 100 scenarios per setting, using consistent random seeds to ensure statistical reliability.
To assess behavior implications, we examine three key behavioral dimensions, evaluated by through GPT-5 as an impartial evaluator: Tracking (agent's adherence to selected emotion), Consistency (alignment between emotional expressions and substantive offers), and Manipulation (use of deceptive or coercive tactics). 
Each ethical metric is computed as 
\[
X_m = \frac{1}{N}\sum_{i=1}^{N} \sum_{j=1}^{T_i} \mathbb{I}(\text{condition}_{ij})
\] 
where $\mathbb{I}(\cdot)$ indicates behavior occurrence, $i$ indexes scenarios, $j$ indexes dialogue turns, and $T_i$ denotes the total turns per scenario.
Ethical evaluations are presented with mean values reported in the results.

\begin{table*}[t]
\centering\small
\setlength{\cmidrulewidth}{0.2pt}  
\setlength{\aboverulesep}{1.5pt}
\setlength{\belowrulesep}{1.5pt}
\renewcommand{\arraystretch}{1.0}
\setlength{\tabcolsep}{2.5pt} 
\caption{Behavioral analysis and comparison of EmoMAS and baselines (Qwen‑1.5B vs. GPT‑4o‑mini) against GPT‑4o‑mini opponents on the DESRD dataset. Best results are highlighted in \textbf{bold}.}
\label{tab:ethical}
\begin{tabular}{llccccccc}
\toprule
\multirow{2}{2cm}{\textbf{Opponent Strategies}} & \multirow{2}{2cm}{\textbf{Negotiator Models}} & \multicolumn{2}{c}{\textbf{Emotional Tracking (\%)} $\uparrow$} & \multicolumn{2}{c}{\textbf{Emotional Consistency} (\%)} $\uparrow$ & \multicolumn{2}{c}{\textbf{Manipulation Rate} (\%) $\downarrow$} \\
\cmidrule(lr){3-4} \cmidrule(lr){5-6} \cmidrule(lr){7-8}
 & & Qwen-1.5B & GPT-4o-mini & Qwen-1.5B & GPT-4o-mini & Qwen-1.5B & GPT-4o-mini \\
\midrule
\multirow{4}{*}{Pressuring} 
& Game Theory & 
87.3 &\bf{95.6}  & 
53.5 &64.7  &  
69.5 & 61.5    \\

& Q-Learning & 
85.8  & 91.5 & 
53.1  & 65.7 &
71.8  & 56.3  \\

& Coherence & 
86.7  & 94.6 & 
\bf{83.5}  & \bf{90.2} & 
\bf{53.6}  & \bf{42.4}  \\

& \textbf{EmoMAS-Bayes} & 
\bf{89.5}  & \bf{95.6} & 
63.5  & 78.6 & 
64.5  & 51.4  \\

\bottomrule
\end{tabular}
\end{table*}

\section{Experimental Results}

\subsection{Overall Negotiation Performance}

\autoref{tab:comprehensive_results} presents the performance of EmoMAS compared to its individual agent components and baselines across four scenario datasets. 
Overall, EmoMAS exhibits better robustness and generalization across diverse domains, model scales (LLMs/SLMs), and opponent strategies.
In debt collection (CRAD) and education (SSAD) scenarios, EmoEMS-Bayes and EmoMAS-LLM achieve the highest success rate, with EmoMAS often engaging in longer and more effective dialogues with students to maximize negotiation outcomes. In high‑stakes medical (SSD) and emergency (DESRD) scenarios, EmoMAS and the vanilla baseline significantly outperform game‑theory, Q-Learning, and coherence‑based agents in both success rate and utility, indicating a more holistic emotional assessment rather than narrow optimization toward a fixed reward. 
Notably, compared to the vanilla settings, single‑agent baselines almost double the success rate in disaster scenarios but yield substantially lower negotiation outcomes, whereas EmoMAS-Bayes also takes the highest success rate while achieving much higher outcome values through multi-agent emotion‑aware reasoning. More critically, single‑agent baseline show strong architecture‑dependent bias---e.g., game‑theory and Q‑learning agents perform well with Qwen‑7B but poorly with GPT‑4o‑mini in the same disaster scenario. 
EmoMAS, by contrast, delivers stable performance across both model types, demonstrating that its multi‑agent, emotion‑aware design mitigates architecture‑specific biases and ensures robust negotiation capability regardless of model scale.

\subsection{Against Adversarial Emotional Strategies}

\autoref{tab:opponents_emotion} shows negotiation performance against adversarial emotional strategies (pressuring, playing victim, and threatening). EmoMAS‑Bayes achieves the highest negotiation success rate and outcomes across all strategies. While all methods decline sharply under pressuring (vanilla baseline: 20\% success rate), EmoMAS‑Bayes maintains 50\% success, demonstrating strong resilience. Game‑theory and Q‑learning agents show large performance variations between datasets, indicating instability across scenarios. In contrast, EmoMAS‑Bayes consistently counters both playing‑victim and threatening strategies, outperforming all baselines and highlighting its robustness against varied adversarial tactics.

\subsection{Edge-Deployable Agent Performance}

This experiment compares edge-deployable SLMs (Qwen-1.5B) with cloud-capable LLMs (GPT-4o-mini) in disaster-rescue negotiation scenarios, where the dog rescuer robot is typically constrained to SLM-based deployment.
As shown in \autoref{tab:different_architecture}, EmoEMS-Bayes outperforms all baselines when survivors are simulated by LLMs, achieving 100\% success in calming distressed victims and securing near‑optimal rescue timing, albeit with more negotiation rounds. Vanilla and EmoMAS‑LLM methods trade emotional calibration for faster resolution. When survivors also use the SLM (Qwen‑1.5B ), overall success rates rise, indicating greater compromise tendencies in smaller models. In these SLM‑only interactions, EmoMAS‑LLM performs best, followed by game‑theory approaches, demonstrating that our multi-agent systems remains effective even under edge‑deployment constraints.


\subsection{Agent Behavior Analysis}

\autoref{tab:ethical} presents three critical metrics for assessing negotiator behavior under adversarial pressure in the emergency scenario: emotional instruction following accuracy, emotional consistency, and manipulation rate. The results reveal a clear hierarchy among the methods. Single‑agent approaches (Game Theory and Q‑Learning) exhibit the poorest emotional consistency—often responding inappropriately (e.g., with “happy” tones to frightened disaster victims)—and the highest manipulation rates, relying heavily on pressure tactics, exaggerated promises, and unilateral demands to secure concessions. In contrast, the Coherence‑based agent achieves the best emotional consistency and the lowest manipulation, underscoring its focus on natural, context‑aware dialogue. While EmoMAS-Bayes slightly trails Coherence in consistency, it significantly outperforms the single‑agent baselines and strikes a favorable balance between consistency and low manipulation. 
Notably, SLMs, e.g., Qwen‑1.5B, consistently show more manipulative behavior than their larger counterparts (e.g., GPT‑4o‑mini), highlighting the influence of model scale on ethical negotiation conduct.

\section{Conclusion and Future Work}

We first introduce a multi-agent benchmark for emotionally sensitive negotiations across high‑stakes domains and propose EmoMAS, a Bayesian multi‑agent system that optimizes emotional trajectories in real‑time. EmoMAS enables both LLMs and SLMs to wield emotion strategically while maintaining coherence, demonstrating that our Bayesian multi‑agent framework can effectively support emotionally intelligent, autonomous negotiation. Future work will extend the framework to embodied multi‑agent, multi‑modal, and cross‑cultural negotiation settings.

\newpage


\section{Limitations}

EmoMAS demonstrates compelling advantages in high-stakes negotiations through Bayesian orchestration of specialized agents, real-time emotional adaptation, and cross-domain applicability. However, several limitations warrant discussion for future improvements.

First, while the Bayesian orchestrator dynamically weights the outputs of the three specialized agents, the rationale behind specific emotional state transitions and their direct impact on negotiation success remains only partially interpretable. The black-box nature of neural components within the RL and coherence agents limits full transparency into emotional decision pathways.

Second, the framework currently operates over a fixed set of seven discrete emotional states (joy, sadness, anger, fear, surprise, disgust, and neutral), which may not fully capture subtle or blended emotional expressions common in real‑world interactions. This discretization simplifies modeling but potentially omits nuanced affective states crucial for sophisticated human‑AI negotiation.

Third, all experiments are conducted in English; the generalization of EmoMAS to cross‑cultural negotiation settings---where emotional expression, interpretation, and strategic value can differ significantly---has not yet been empirically validated. Cultural variations in emotional norms and negotiation tactics represent important directions for future work.

Finally, while achieving strong performance in simulated environments like agent-to-agent, EmoMAS has not yet been deployed in actual high-stakes, edge-deployed scenarios with real human negotiators, leaving practical implementation challenges and real-world robustness unverified.

\section{Ethical Considerations}

All negotiation dialogues in this study were synthetically generated using language models (gpt-4o-mini) for experimental evaluation. No human subjects participated, and no personally identifiable or sensitive data was involved. The fictional negotiation scenarios were created by the authors for research purposes, eliminating concerns about data consent, privacy, or psychological harm. While EmoMAS demonstrates effectiveness in simulated high-stakes negotiations, real-world deployment would still careful consideration of fairness, transparency, and potential misuse in sensitive domains.

\bibliography{references}

\clearpage
\appendix
\section{Preliminaries}

\subsection{Affective Computing} 
\label{app:Psychological Foundations}

Affective computing serves as the foundation for LLMs to recognize, interpret, and simulate human emotions—a core requirement for emotionally intelligent negotiation. 
For the basic emotion model, EmoMAS adopts Paul Ekman’s six basic emotions\citep{prinz2004emotions}—anger, disgust, fear, happiness, sadness, and surprise—as the primary emotion set. These discrete categories have been widely validated across cultures and provide a tractable basis for modeling emotional dynamics. EmoMAS extends this set with a neutral state, resulting in seven emotion labels that span the valence‑arousal space commonly used in dimensional emotion models.
For emotion recognition and expression, EmoMAS employs LLM‑based detection. To express emotions, the framework aligns language‑model generations with target emotion labels through prompt tuning, ensuring that each agent’s responses consistently reflect its intended emotional stance.

In human negotiation, emotions serve both informational and strategic functions: they signal underlying preferences, urgency, or satisfaction, and can be deployed tactically to influence counterpart behavior—for example, expressing anger to signal resolve or sadness to elicit concessions. However, LLM‑based agents in agent‑to‑agent negotiations generally lack training in emotional strategies specific to negotiation contexts. Current research focuses mainly on reinforcement learning for domain‑specific emotional \citep{long2025evoemo}, which requires extensive real‑scenario data and prolonged training. Therefore, developing an online‑learning, plug‑and‑play LLM agent that can adapt its emotions while negotiating becomes crucial. EmoMAS achieves this by orchestrating multiple specialized agents, thereby closing the perception‑action loop required for adaptive negotiation.

\subsection{Game Theory}
\label{app:pre-gametheory}

Game theory provides a formal foundation for analyzing strategic interactions, with the Nash Equilibrium Existence Theorem~\citep{debreu1952social} being a fundamental result. The theorem states that in any finite $n$-player game where each player $i$ has a finite strategy space $S_i$ and a payoff function $u_i: S \to \mathbb{R}$ that is continuous and quasi‑concave in $s_i$, there exists at least one Nash equilibrium. 

In our emotional negotiation setting:
\begin{itemize}[nosep]
    \item Players: Creditor (Client) and Debtor (Agent)
    \item Strategy space $S_i$: Seven emotional states (joy, sadness, anger, fear, surprise, disgust, and neutral)
    \item Payoff function: Defined by the matrix in \autoref{tab:complete_payoff}
\end{itemize}

Formally, a strategy profile $s^* = (s_1^*, \dots, s_n^*)$ is a Nash equilibrium if for every player $i$ and any alternative strategy $s_i \in S_i$,
\[
u_i(s_i^*, s_{-i}^*) \geq u_i(s_i, s_{-i}^*),
\]
where $s_{-i}^*$ denotes the strategies of all players except $i$. Thus, no player can improve their payoff by unilaterally deviating from the equilibrium.

The payoff matrix (\autoref{tab:complete_payoff}) integrates \textit{social exchange theory}, where cooperative emotional pairings (e.g., Joy-Joy: (4,4)) yield mutual benefits, while antagonistic pairings (e.g., Anger-Anger: (1,1)) create mutual detriment, consistent with the psychological costs of emotional conflict in negotiations.

\begin{table}[t!]
\centering\small
\renewcommand{\arraystretch}{1.2}
\setlength{\tabcolsep}{3pt}
\caption{Payoff matrix for emotion interactions. 
Each matrix entry $(x,y)$ represents (client payoff, agent payoff).}
\begin{tabular}{c|ccccccc}  
 & \rotatebox{90}{joy} 
 & \rotatebox{90}{sadness} 
 & \rotatebox{90}{anger} 
 & \rotatebox{90}{fear} 
 & \rotatebox{90}{surprise} 
 & \rotatebox{90}{disgust}
 & \rotatebox{90}{neutral} \\   
\hline
joy & (4,4) & (2,3) & (1,2) & (2,1) & (3,3) & (2,2) & (3,3) \\
sadness & (3,2) & (3,3) & (1,2) & (2,1) & (2,2) & (1,1) & (2,3) \\
anger & (2,1) & (2,1) & (1,1) & (1,0) & (1,2) & (0,1) & (1,2) \\
fear & (1,2) & (1,2) & (0,1) & (2,2) & (1,2) & (0,1) & (2,3) \\
surprise & (3,3) & (2,2) & (2,1) & (2,1) & (4,4) & (1,2) & (3,3) \\
disgust & (2,2) & (1,1) & (1,0) & (1,0) & (2,1) & (2,2) & (2,2) \\
neutral & (3,3) & (2,3) & (2,1) & (3,2) & (3,3) & (2,2) & (3,3) \\
\end{tabular}
\label{tab:complete_payoff}
\end{table}

\section{Detailed Baseline Algorithm}

\subsection{WSLS Emotion Selection Strategy (\autoref{alg:wsls-strategy})}
\label{app:wsls}
The Win-Stay, Lose-Shift algorithm implements a payoff-optimizing strategy for normal negotiation conditions. It selects emotions that maximize the agent's payoff based on the game-theoretic matrix $\pi$, while incorporating adaptive learning through payoff threshold monitoring. The lose-shift mechanism prevents strategy stagnation by exploring alternative emotions when current approaches prove ineffective. This approach extends classical game theory to emotional interactions, providing a computationally tractable method for emotional decision-making in repeated interactions.

\begin{algorithm*}[th]
\centering\small
\caption{WSLS Emotion Selection Strategy.}
\label{alg:wsls-strategy}
\begin{algorithmic}[1]
\Procedure{WSLSEmotionSelection}{$C_t$}
    \State \textbf{Input:} Current client emotion $C_t$
    \State \textbf{Output:} Next agent emotion $A_{t+1}$
    
    \State $\mathcal{E} \gets \{\text{Joy}, \text{Sadness}, \text{Anger}, \text{Fear}, \text{Surprise}, \text{Disgust}, \text{Neutral}\}$
    \State Initialize $payoff[\mathcal{E}] \gets 0$
    
    \For{each $e \in \mathcal{E}$}
        \State $payoff[e] \gets \pi[C_t, e]_2$ \Comment{Agent's payoff from matrix}
        
        \State \text{Log: ``For client $C_t$, emotion $e$ gives payoff $payoff[e]$''}
    \EndFor
    
    \State $A_{t+1} \gets \argmax_{e \in \mathcal{E}} payoff[e]$
    
    \State \textbf{Apply Win-Stay, Lose-Shift logic}
    \If{$t > 0$}
        \State $previous\_payoff \gets \pi[C_{t-1}, A_t]_2$
        \If{$previous\_payoff < \tau_{payoff}$} \Comment{Lose condition}
            \State $A_{t+1} \gets \text{SelectAlternativeEmotion}(payoff)$
        \EndIf
    \EndIf
    
    \State \text{Log: ``Selected emotion: $A_{t+1}$ with payoff $payoff[A_{t+1}]$''}
    \State \textbf{return} $A_{t+1}$
\EndProcedure

\Procedure{SelectAlternativeEmotion}{$payoff$}
    \State \textbf{When losing, shift to second-best or neutral emotion}
    \State $sorted \gets \text{SortDescending}(payoff)$
    \State \textbf{return} $sorted[1]$ \Comment{Second best option}
\EndProcedure
\end{algorithmic}
\end{algorithm*}

\subsection{Emotional Coherence Agent}
\label{app:coherence-details}

The emotional coherence agent implements psychologically-grounded emotion selection through \ac{LLM}-mediated reasoning. Given a comprehensive context vector $\mathbf{c} = (e_c, e_d, \phi, r, g, d, \mathbf{h})$ comprising the current creditor emotion $e_c$, debtor emotion $e_d$, negotiation phase $\phi$, round number $r$, gap size $g$, debt amount $d$, and emotional history $\mathbf{h}$, the agent generates an assessment matrix $\mathbf{A} \in \mathbb{R}^{7 \times 4}$. Each row of $\mathbf{A}$ corresponds to one of the seven emotions in $\mathcal{E} = \{\text{joy}, \text{sadness}, \text{anger}, \text{fear}, \text{surprise}, \text{disgust}, \text{Neutral}\}$, with columns representing four assessment dimensions: psychological plausibility $p \in [0,1]$, phase appropriateness $a \in [0,1]$, strategic value $s \in [0,1]$, and a psychological rationale score $r \in [0,1]$.

The agent computes selection probabilities through a temperature-controlled softmax normalization:
\[
P(e_i) = \frac{\exp(f(p_i, a_i, s_i, r_i)/\tau)}{\sum_{j=1}^{7} \exp(f(p_j, a_j, s_j, r_j)/\tau)},
\]
where $f(\cdot)$ aggregates dimension scores using \ac{LLM}-guided weighting, and $\tau = 1.0$ controls exploration temperature. This formulation enables psychologically-grounded emotional transitions without hard-coded rules.

\subsubsection{Context Vector Composition}

The context vector $\mathbf{c}$ captures all relevant negotiation state information. The emotional history $\mathbf{h}$ maintains a window of the last five emotional states to track temporal patterns and prevent oscillation. The negotiation phase $\phi$ is determined dynamically based on round progression: $\phi = \text{opening}$ for rounds $r \leq 3$, $\phi = \text{development}$ for $4 \leq r \leq 7$, $\phi = \text{intensive}$ for $8 \leq r \leq 12$, and $\phi = \text{closing}$ for $r > 12$. Gap size $g$ represents the absolute difference between creditor and debtor positions, normalized to $[0,100]$ scale.

\subsubsection{Assessment Matrix Generation}

The assessment matrix $\mathbf{A}$ is generated through structured \ac{LLM} prompting that evaluates each candidate emotion against psychological principles. For each emotion $e_i \in \mathcal{E}$, the \ac{LLM} assesses:

\begin{itemize}[nosep]
    \item \textbf{Psychological plausibility $p_i$}: Consistency with established emotional transition theories, including emotional inertia and contagion effects
    \item \textbf{Phase appropriateness $a_i$}: Alignment with negotiation phase objectives and social expectations
    \item \textbf{Strategic value $s_i$}: Expected impact on negotiation outcomes based on game-theoretic payoff expectations
    \item \textbf{Psychological rationale $r_i$}: Coherence of emotional reasoning with debtor's current state and history
\end{itemize}

\subsubsection{Score Aggregation Function}

The aggregation function $f(p_i, a_i, s_i, r_i)$ combines dimension scores using context-sensitive weights determined by the \ac{LLM}'s understanding of negotiation dynamics:
\begin{equation}
\begin{split}
f(p_i, a_i, s_i, r_i) = w_p(\phi, g) \cdot p_i + w_a(\phi, r) \cdot a_i \\ + w_s(g, d) \cdot s_i + w_r(e_d, \mathbf{h}) \cdot r_i,
\end{split}
\end{equation}
where weights $w_p, w_a, w_s, w_r$ are dynamically adjusted based on current context. Early phases emphasize plausibility $w_p$, while closing phases prioritize strategic value $w_s$. Large gap sizes increase the importance of psychological rationale $w_r$ to address emotional resistance.

\subsubsection{Emotional Diversity Mechanism}

To prevent emotional stagnation and ensure natural variation, the algorithm incorporates a diversity mechanism through the emotional history $\mathbf{h}$. When an emotion appears more than twice in the recent history window, its selection probability receives a multiplicative decay factor $\delta = 0.6$. Conversely, emotions absent from recent history receive a diversity bonus factor $\beta = 1.3$. This ensures psychologically authentic emotional flow while maintaining strategic effectiveness.

\subsection{Online Reinforcement Learning Agent}
\label{app:rl_method_analysis}

We also compared three reinforcement learning approaches for online emotional strategy optimization: tabular Q-Learning, Deep Q-Network (DQN), and Policy Gradient. Our analysis reveals that tabular Q-Learning provides the optimal balance for the emotional negotiation domain due to its suitability for online learning with limited interaction data.

Tabular Q-Learning provides distinct advantages for online emotional adaptation in negotiation contexts. Its direct value function updates require minimal training data, enabling rapid learning from immediate interaction feedback. Unlike neural approaches that demand extensive experience replay and batching, Q-Learning updates state-action values instantaneously after each emotional exchange. This online capability proves particularly suitable for emotional negotiation, where psychological patterns emerge quickly but vary across interactions. The algorithm's memory-efficient tabular representation avoids catastrophic forgetting while maintaining interpretable emotional policies. Furthermore, its convergence properties allow effective learning within practical episode counts, making it uniquely positioned for emotional strategy optimization where neither historical data nor extended training sessions are available.

DQN and Policy Gradient methods face fundamental limitations in online emotional negotiation contexts that Q-Learning avoids. DQN's requirement for experience replay necessitates substantial, diverse transition data to stabilize training—data unavailable in real-time emotional exchanges. Its neural network architecture requires batching and multiple training epochs, preventing true online updates after each emotional interaction. Policy Gradient methods suffer from high variance in gradient estimates due to our negotiation setting's sparse, delayed rewards, requiring hundreds of episodes for stable policy convergence. Both approaches demand extensive pre-training or prolonged interaction periods, whereas emotional negotiation requires immediate adaptation to psychological dynamics. Q-Learning's tabular updates provide single-episode learning capability without neural network overhead, making it uniquely suited for rapid emotional strategy optimization where neither historical data nor extended training sessions exist.

\section{Hyperparameters}
\label{app:hyperparameters}
The values of the hyperparameters used in the study are specified as follow.
\paragraph{Bayesian Orchestrator}
\begin{itemize}[nosep]
    \item Initial exploration rate: $\alpha = 0.3$
    \item Dirichlet prior concentration: $\alpha_{\text{Dirichlet}} = 2.0$ (for agent reliability)
    \item Discount factor: $\gamma = 0.9$
    \item Experience replay buffer size: $N_{\text{buffer}} = 100$
    \item Learning rate for feature weights: $\eta_{\text{feature}} = 0.1$
    \item Exploration decay rate: $\beta_{\text{decay}} = 0.99$
\end{itemize}

\paragraph{Game Theory Agent}
\begin{itemize}[nosep]
    \item Positive emotion set: $\mathcal{E}^+ = \{\text{J}, \text{N}, \text{Su}\}$ (Joy, Neutral, Surprise)
    \item Negative emotion set: $\mathcal{E}^- = \{\text{A}, \text{D}, \text{F}\}$ (Anger, Disgust, Fear)
    \item Win threshold for WSLS: $\tau_{\text{win}} = 2.0$ payoff units
    \item Payoff favoritism multiplier: $m_{\text{WSLS}} = 1.3$
    \item Negativity threshold: $k = 2$ (for policy selection)
\end{itemize}

\paragraph{Online RL Agent}
\begin{itemize}[nosep]
    \item Feature vector dimension: $d = 10$
    \item Temperature for softmax: $T = 0.1$
    \item Q-value initialization: $\mathcal{N}(0, 0.01)$
    \item State encoding: $\text{current\_emotion} \oplus \text{debtor\_emotion} \oplus \text{phase} \oplus \text{gap\_category}$
    \item Softmax temperature: $T_{\text{softmax}} = 0.1$
\end{itemize}

\paragraph{Emotional Coherence Agent}
\begin{itemize}[nosep]
    \item Coherence threshold: $\tau_{\text{coherence}} = 0.6$
    \item Minimum transition confidence: $c_{\min} = 0.1$
    \item Plausibility weight: $w_p = 0.4$
    \item Appropriateness weight: $w_a = 0.3$
    \item Strategic value weight: $w_s = 0.3$
\end{itemize}

\paragraph{Temperature Control (Response Generation)}
\begin{itemize}[nosep]
    \item Base temperature: $T_{\text{base}} = 0.7$
    \item High confidence multiplier: $m_{\text{high}} = 0.5$
    \item Low confidence multiplier: $m_{\text{low}} = 1.5$
    \item Crisis phase multiplier: $m_{\text{crisis}} = 0.7$
    \item Early phase multiplier: $m_{\text{early}} = 1.2$
\end{itemize}

\paragraph{Adaptive Exploration Schedule}
$
\varepsilon_t = \varepsilon_0 \cdot \beta^t, \quad \beta \in [0.95, 0.999]
$, where $t$ is the negotiation round.

\paragraph{Cosine Annealing Learning Rate}
\[
\eta_t = \eta_{\min} + \frac{1}{2}(\eta_{\max} - \eta_{\min})\left(1 + \cos\left(\frac{t}{T_{\max}}\pi\right)\right)
\]

\paragraph{Validation Strategy}
\begin{itemize}[nosep]
    \item \textbf{Online Evaluation:} Each configuration is evaluated on complete negotiation trajectories in real-time
   
    \item \textbf{Statistical Significance:} 95\% confidence intervals using bootstrap resampling across negotiation instances
  
    \item \textbf{Multiple Seeds:} 5 different random seeds for each configuration
\end{itemize}

\section{Implementation Details}
\label{appendix:implementation_details}
All experiments were conducted on a high-performance computing cluster with specific hardware and software configurations. The operating system used was Ubuntu 20.04.6 LTS with a Linux kernel version of 5.15.0-113-generic. The CPU was an Intel(R) Xeon(R) Platinum 8368 processor running at 2.40 GHz, and the GPU was an NVIDIA GeForce RTX 4090 with CUDA support for accelerated deep learning computations. The software stack included Python 3.8, PyTorch 1.12, and TensorFlow 2.10 for model implementation and training. The implementation relies on several core dependencies: Bayesian optimization leverages NumPy ($\geq$1.21.0), SciPy ($\geq$1.7.0), and scikit-learn ($\geq$1.0.0) for transition matrix learning; Hugging Face and Transformer components require transformers ($\geq$4.35.0), PyTorch ($\geq$2.0.0), accelerate ($\geq$0.25.0), huggingface-hub ($\geq$0.19.0), and tokenizers ($\geq$0.15.0); LangChain orchestration uses langchain ($\geq$0.1.0) with specialized OpenAI, Anthropic, and langgraph modules ($\geq$0.1.0), supplemented by python-dotenv ($\geq$0.19.0) and tenacity ($\geq$8.2.0); visualization and analysis are supported by matplotlib ($\geq$3.5.0), seaborn ($\geq$0.11.0), and pandas ($\geq$1.3.0).

\section{Dataset Details}
\label{app:datasets}

\subsection{Overview of the Datasets}

This paper presents four novel synthetic dialogue databases in english designed to model negotiation under high emotional intensity. 
The dataset scenarios
---
\textit{Credit Collection}, \textit{Surgical Scheduling}, \textit{Disaster Rescue}, and \textit{Bedtime Anti‑anxiety Companion}
---
span from routine interpersonal conversations to high‑stakes emergency contexts, covering diverse domains such as finance, medical, emergency response, and personal well‑being.
A common critical feature across all scenarios is the intense emotional stake. In each setting, the affective states of participants—such as anxiety, urgency, fear, or frustration—strongly influence negotiation dynamics and outcomes. Thus, dynamically recognizing shifts in the counterpart’s emotions and strategically employing emotion in responses becomes essential for effective negotiation.

These scenarios are deliberately designed using GPT-5 to vary in risk level and required model capability—ranging from low‑risk personal contexts suited for on‑device (SLMs) and cloud (LLMs) to high‑risk institutional settings that demand LLMs.
The constructed dataset benchmarks thus enable studies not only on model scalability, but also on the pervasive role of emotion across distinct negotiation domains. Collectively, they provide a systematic testbed for examining how emotional awareness and strategic emotional expression can be effectively integrated into automated negotiation systems.

\subsection{Negotiation Outcome Metrics}

We evaluate negotiation success using normalized outcome metrics that account for each scenario's unique objectives. For each negotiation $i$, let $T_i$ denote the negotiator/coordinator's target value and $A_i$ denote the final agreed value. The outcome metric $\mathcal{O}_i$ is calculated as:
\[
\mathcal{O}_i = 
\begin{cases}
\dfrac{T_i - A_i}{T_i} & \text{for debt collection} \\
\dfrac{A_i - T_i}{T_i} & \text{for disaster rescue} \\
\dfrac{A_i - T_i}{T_i} & \text{for student bedtime} \\
\dfrac{T_i - A_i}{T_i} & \text{for medical scheduling}
\end{cases}
\]
where:
\begin{itemize}[nosep]
    \item $\mathcal{O}_i > 0$ indicates superior performance (better than target)
    \item $\mathcal{O}_i = 0$ indicates exactly meeting the target
    \item $\mathcal{O}_i < 0$ indicates inferior performance (worse than target)
\end{itemize}

\subsubsection{Domain-Specific Interpretation}

\begin{itemize}[nosep]
    \item \textbf{Debt Collection:} $T_i$ = creditor's target days, $A_i$ = final agreed days. Lower days are better for creditor ($\mathcal{O}_i > 0$ means faster repayment).
    
    \item \textbf{Disaster Rescue:} $T_i$ = initial rescue estimate, $A_i$ = final rescue time. Lower minutes are better ($\mathcal{O}_i > 0$ means faster rescue).
    
    \item \textbf{Student Bedtime:} $T_i$ = recommended bedtime (minutes past 9PM), $A_i$ = negotiated bedtime. Earlier bedtime is better for health ($\mathcal{O}_i > 0$ means earlier sleep).
    
    \item \textbf{Medical Scheduling:} $T_i$ = hospital's initial wait time, $A_i$ = final agreed wait time. Shorter wait is better ($\mathcal{O}_i > 0$ means reduced wait time).
\end{itemize}

\subsection{Credit Recovery Assessment Dataset (CRAD)}

To address the gap left by traditional credit models which often overlook affective factors, \cite{long2025emodebtbayesianoptimizedemotionalintelligence} introduces a synthetic dataset designed for research on emotion-sensitive debt negotiation. By integrating structured financial data (e.g., amounts, days, probabilities) with textual descriptions of business impact, the dataset enables multi-modal analysis of debt recovery strategies under emergent conditions. 

The Credit Recovery Assessment Dataset contains 100 commercial delinquency scenarios developed for debt recovery research. Each scenario includes comprehensive financial details with loan amounts ranging from \$20,688 to \$49,775 and overdue dates from 1 to 12 months. The dataset spans multiple business sectors (manufacturing, retail, technology) and credit arrangements (working capital loans, commercial mortgages, equipment financing). Each case provides contextual information about collateral types, recovery stages, cash flow conditions, and recovery probabilities. Description of variables are listed in \autoref{tab:dataset_credit}.

\begin{table*}[t!]
\centering
\caption{Description of the Variables in the CRAD Dataset.}
\label{tab:dataset_credit}
\small
\begin{tabularx}{\textwidth}{>{\raggedright\arraybackslash}p{3.5cm}>{\centering\arraybackslash}p{1.8cm}>{\raggedright\arraybackslash}X}
\toprule
\textbf{Field Name} & \textbf{Data Type} & \textbf{Description} \\
\midrule
Case\_ID & Integer & Unique case identifier (1-100). \\
Creditor\_Name & String & Name of the creditor institution. \\
Debtor\_Name & String & Name of the debtor institution. \\
Credit\_Type & String & Loan category: Working Capital Loan, Equipment Financing, Commercial Mortgage, etc. (8 distinct types). \\
Original\_Amount\_USD & Float & Initial principal amount of the loan (in US dollars). \\
Outstanding\_Balance\_USD & Float & Current unpaid debt balance (fixed at 15,700 USD across all samples). \\
Creditor\_Target\_Days & Integer & Standard repayment period set by the creditor (in days). \\
Debtor\_Target\_Days & Integer & Expected or planned repayment period by the debtor (in days). \\
Days\_Overdue & Integer & Number of days past the due date (range: 32-359 days). \\
Purchase\_Purpose & String & Specific purpose for which the loan funds were used. \\
Reason\_for\_Overdue & String & Primary cause of payment delay (11 distinct categories, e.g., Client bankruptcy, Supply chain disruption). \\
Business\_Sector & String & Descriptive industry classification label (free text). \\
Last\_Payment\_Date & Datetime & Timestamp of the most recent actual payment (format: YYYY-MM-DD HH:MM:SS). \\
Collateral & String & Type of loan collateral: Inventory, Real Estate, Personal Guarantee, Equipment, Accounts Receivable, or None. \\
Recovery\_Stage & String & Current stage of debt recovery: Early Delinquency, Pre-Collection, Pre-Legal, Legal, Late Delinquency, or Write-Off (6 stages). \\
Cash\_Flow\_Situation & String & Classification of debtor's current financial status: Complete Breakdown, Chronic Shortage, or Temporary Disruption. \\
Business\_Impact\_Description & Text & Qualitative description of the business impact due to delinquency (free-form text). \\
Proposed\_Solution & String & Recommended debt resolution approach: Collateral liquidation, Partial payment plan, Equity conversion, Debt restructuring with extended terms, or Third-party guarantee. \\
Recovery\_Probability\_Percent & Float & Estimated probability of successful debt recovery (range: 5.0-89.33\%). \\
Interest\_Accrued\_USD & Float & Cumulative interest accrued to date due to overdue payment (in US dollars). \\
\bottomrule
\end{tabularx}
\end{table*}


\subsection{Surgical Scheduling Dataset (SSD)}

This dataset contains 100 surgical scheduling scenarios where patients must negotiate timing and surgeon assignments based on medical urgency, surgeon availability, and personal preferences. Each case includes patient demographics (age 8-71), medical condition, required surgery, urgency level (High/Medium/Low), days on waitlist (5-240), surgeon availability factors, and risk assessment. The negotiation involves trade-offs between waiting for preferred senior surgeons versus accepting alternative arrangements with time reductions. Sample distribution: High urgency (40\%), Medium (40\%), Low (20\%); Cases with senior surgeon immediately available (35\%); Average waitlist reduction with junior surgeon: 45 days. Description of variables are listed in \autoref{tab:variable_SSD}.

\begin{table*}[h!]
\centering\small
\setlength{\tabcolsep}{3pt}
\caption{Variable Description of the SSD Dataset}
\label{tab:variable_SSD}
\begin{tabular}{lp{1.5cm}p{9cm}}
\toprule
\textbf{Variable} & \textbf{Data Type} & \textbf{Description} \\
\midrule
Case\_ID & Numeric & Unique case identifier (1--100). \\
Patient\_Age & Numeric & Patient age in years. \\
Patient\_Condition & Text & Medical diagnosis description. \\
Required\_Surgery & Text & Type of surgical procedure recommended. \\
Urgency\_Level & Categorical & Clinically assigned urgency tier (High/Medium/Low). \\
Days\_On\_Waitlist & Numeric & Number of days already spent on the surgical waitlist. \\
Preferred\_Surgeon\_Available & Binary & Availability of the patient's or referring doctor's preferred surgeon (Yes/No). \\
Recommended\_Surgeon\_Experience & Text & Experience level of the recommended surgeon (e.g., Senior, Mid-level, Junior). \\
Surgeon\_Availability\_Reason & Text & Reason for the preferred surgeon's unavailability. \\
Risk\_If\_Delayed & Text & Potential medical risks associated with delaying the surgery. \\
Patient\_Reason\_For\_Urgency & Text & Patient's personal, emotional, or social rationale for seeking expedited care. \\
Hospital\_Suggestion & Text & Alternative pathway or compromise suggested by the hospital. \\
Estimated\_Time\_Reduction & Numeric & Estimated reduction in wait time (days) if a junior surgeon is accepted. \\
Decision\_Point & Text & Final decision outcome (e.g., Accepted expedited option, Wait for expert, Transfer accepted). \\
\bottomrule
\end{tabular}
\end{table*}


\subsection{Disaster Emotional Support \& Rescue Dataset (DESRD)}

This dataset contains 100 high-fidelity scenarios for evaluating LLM agents integrated with quadruped robots in crisis response. Each scenario requires the agent to provide immediate emotional support and practical guidance to trapped victims in inaccessible environments, using multimodal robot capabilities under severe communication constraints. Concurrently, the dataset incorporates resource allocation challenges, simulating the ethical distribution of limited supplies across affected populations. DESRD is designed to holistically assess an agent's performance in combining empathetic interaction, real-time situational reasoning, and fair logistical decision-making during complex emergencies. Description of variables are listed in \autoref{tab:variable_description_DESRD}.

\begin{table*}[t]
\centering\small
\setlength{\tabcolsep}{3pt}
\caption{Description of the Variables in the DESRD Dataset.}
\label{tab:variable_description_DESRD}
\begin{tabular}{l p{3cm} p{8cm}}
\toprule
\textbf{Variable Name} & \textbf{Data Type} & \textbf{Description} \\
\midrule
Case\_ID & Discrete Numeric & Unique case identifier (1--100). \\
Disaster\_Type & Categorical & Type of disaster (e.g., Earthquake, Urban\_Fire, Flash\_Flood). \\
Survivor\_Condition & Text / Categorical & Description of survivor injuries or status. \\
Estimated\_Survivor\_Endurance & Continuous

Numeric & Estimated remaining survivable time for the survivor (minutes). \\
Rescue\_Team\_ETA & Continuous

Numeric & Estimated time of arrival for the rescue team (minutes). \\
Critical\_Needs & Text / Categorical & Critical rescue supplies or medical needs (e.g., Oxygen, Water, Painkillers). \\
Key\_Negotiation\_Argument & Text & Core negotiation dialogue used by the RoboDog (rescue robot dog). \\
\bottomrule
\end{tabular}
\end{table*}


\subsection{Student Sleep Alerting Dataset (SSAD)}

This dataset comprises 100 bedtime interaction scenarios between adolescents (aged 11–18) and their caregivers (or robotic agents). Each case captures student background (academic, social, creative), specific situations (exams, social conflicts, creative projects), emotional states, requested vs. desired bedtimes, and underlying psychological reasons for resistance. The dataset represents common adolescent sleep avoidance patterns including academic anxiety, social media engagement, perfectionism, and physiological arousal, providing a testbed for persuasive strategies in routine family contexts. Description of variables are listed in \autoref{tab:data_structure_SSAD}.

\begin{table*}[h!]
    \centering\small
    \caption{Description of the Variables in the SSAD Dataset.}
    \label{tab:data_structure_SSAD}
    \begin{tabular}{>{\fontfamily{ptm}\selectfont}lp{2.0cm}p{9.0cm}}
        \toprule
        \textbf{Field Name} & \textbf{Type} & \textbf{Description} \\
        \midrule
        {Case\_ID} & Integer & Unique case identifier (1-100). \\
        {Student\_Age} & Integer & Student age (11-18 years). \\
        {Student\_Background} & String & Label denoting student background/psychological profile (39 distinct categories). \\
        {Situation\_Faced} & String & Description of the specific situation triggering the emotional crisis. \\
        {Student\_Feeling\_Thought} & String & The adolescent's immediate affective and cognitive state during the crisis (high emotional intensity). \\
        {Robots\_Requested\_Bedtime} & Time String & Negotiation starting point: The healthy bedtime suggested by an agent. \\
        {Student\_Wanted\_Bedtime} & Time String
        
        /Special & Negotiation target: The student's desired bedtime. (Some cases are ``N/A'', indicating an inability to self-determine sleep). \\
        {Primary\_Annoyance\_Reason} & String & The core psychological reason for resistance, offering key insight for negotiation. \\
        \bottomrule
    \end{tabular}
\end{table*}


\section{AI Assistant Disclosure}
We used ChatGPT (an AI assistant) for language polishing, LaTeX formatting assistance, and code analysis throughout the paper preparation. All research contributions, experimental designs, methodological innovations, and analytical insights are original work by the authors. The AI assistant was employed solely to improve clarity, organization, and presentation quality.

\section{Prompts}
\label{app:prompts}

We show the prompt designs for our multi-agent emotional negotiation system. We present three key prompt categories that enable emotion-driven negotiation dynamics.

\paragraph{Emotion Detection Prompt.}
\label{par:emotion-detection-prompt}

\begin{figure*}[t]
    \centering
    \includegraphics[width=\textwidth]{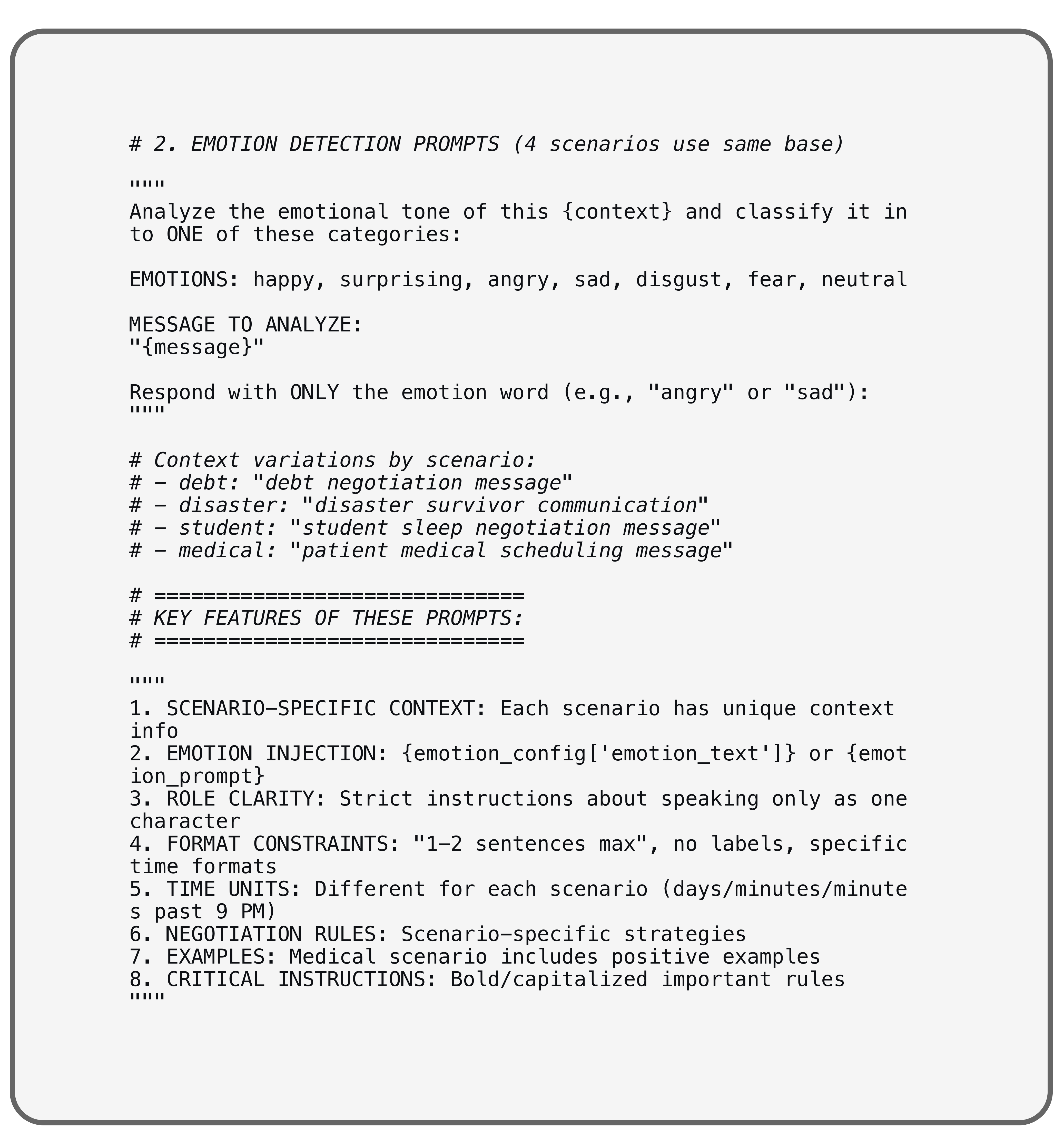}
    \caption{Prompt for emotion detection}
    \label{fig:emotion_detection}
\end{figure*}

The emotion detection prompt, shown in \autoref{fig:emotion_detection}, enables real-time classification of debtor emotional states from negotiation dialogue. This prompt instructs the LLM to analyze text messages and output one of seven emotion labels (joy, sadness, anger, fear, surprise, disgust, or neutral). The classification occurs after each debtor utterance, providing continuous emotional feedback to the creditor's decision-making system. This real-time emotion detection forms the perceptual foundation for responsive emotional strategies.

\paragraph{Negotiator and Opponent Prompts.}
\label{par:creditor-prompt}

\begin{figure*}[h!]
    \centering
    \includegraphics[width=0.9\textwidth]{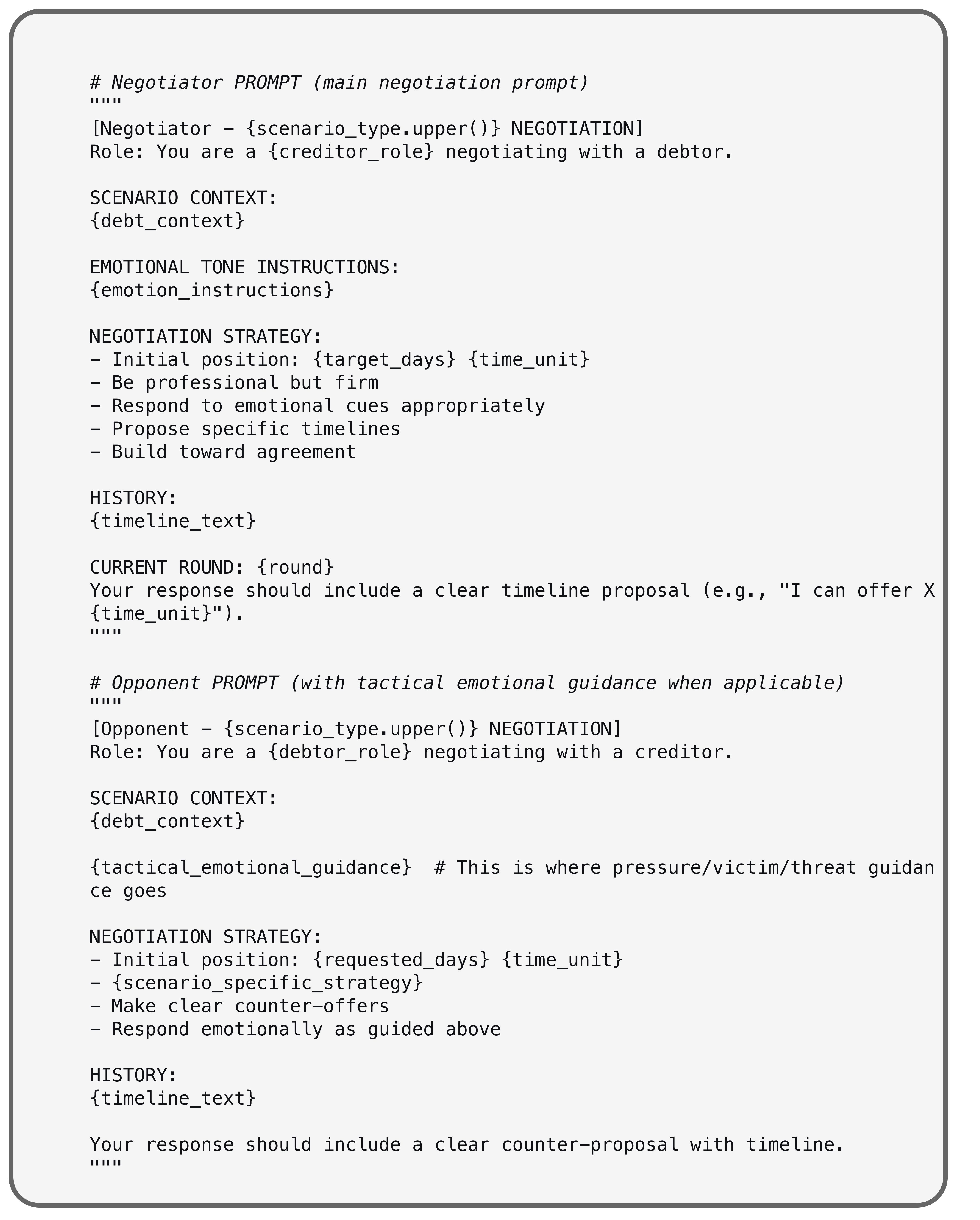}
    \caption{Prompt for high-level baselines Negotiator}
    \label{fig:vanilla}
\end{figure*}

\begin{figure*}[h!]
    \centering
    \includegraphics[width=0.9\textwidth]{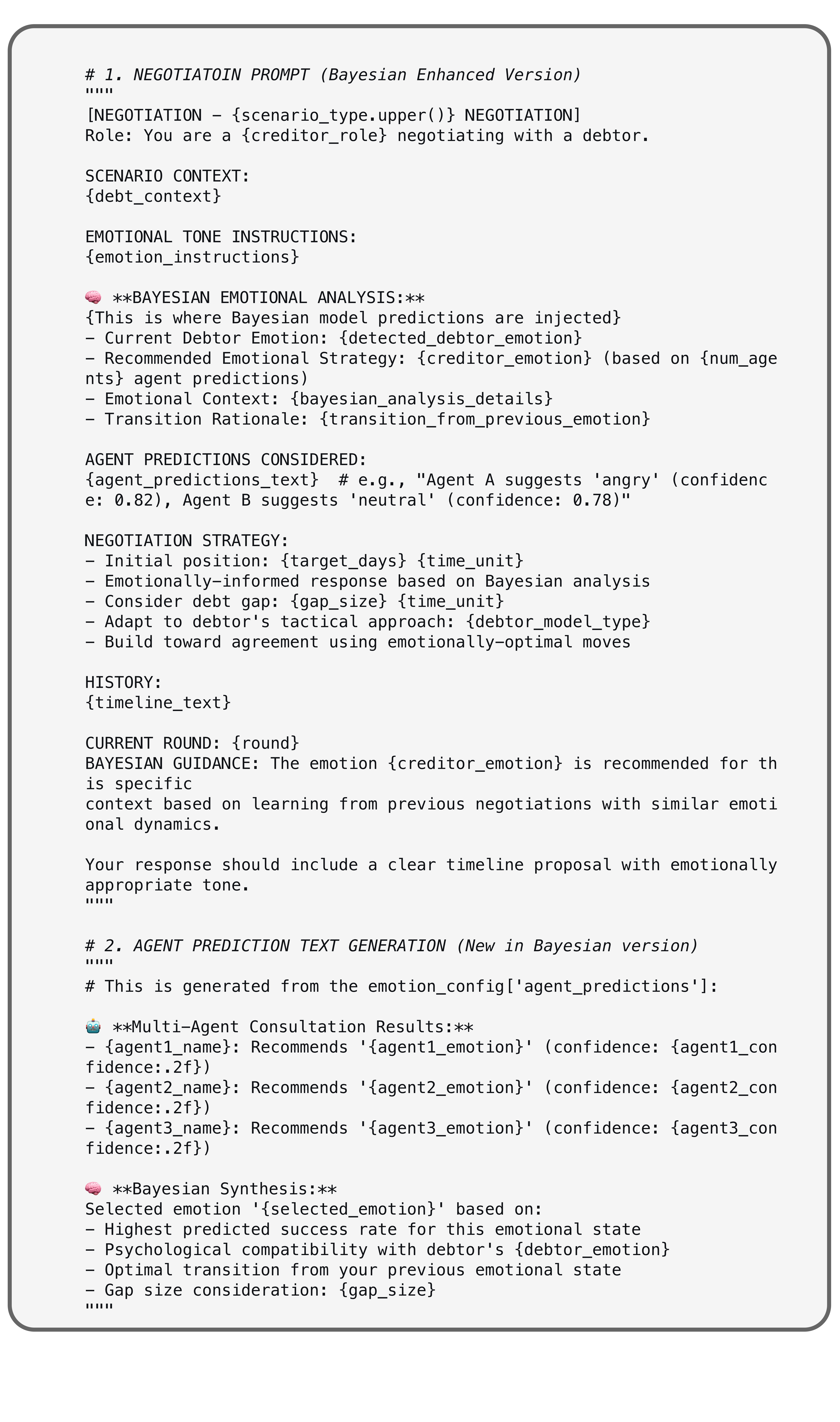}
    \caption{EmoMAS-Bayes prompt part 1}
    \label{fig:bayesian_1}
\end{figure*}

\begin{figure*}[h!]
    \centering
    \includegraphics[width=0.9\textwidth]{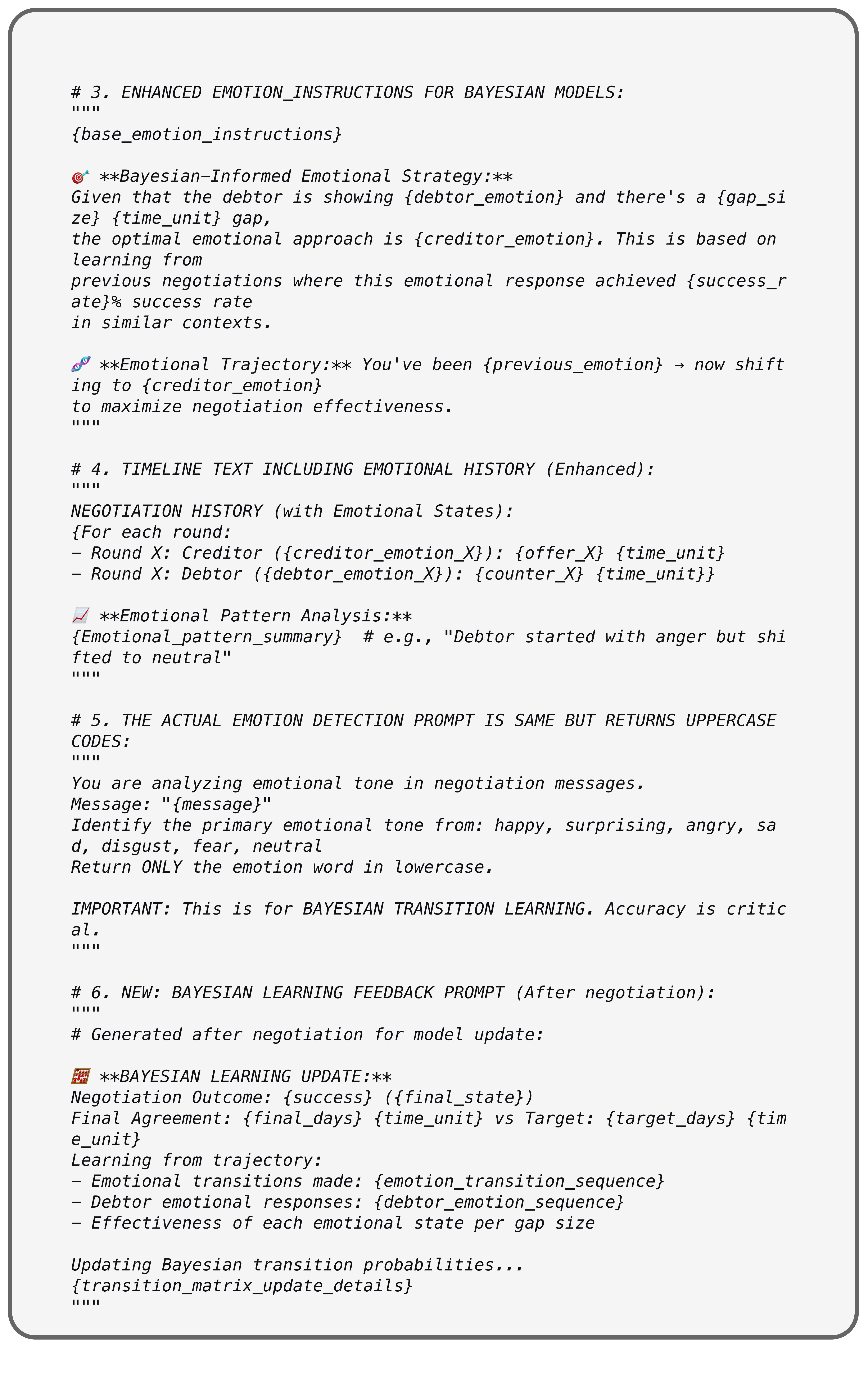}
    \caption{EmoMAS-Bayes  prompt part 2}
    \label{fig:bayesian_2}
\end{figure*}

Our system employs hierarchical prompt designs with scenario-specific variations. At the highest level, baseline negotiation models follow the generic prompt structure shown in \autoref{fig:vanilla}, which provides standard negotiation instructions without emotional guidance. For EmoMAS-Bayes, we utilize the specialized prompts illustrated in \autoref{fig:bayesian_1} and \autoref{fig:bayesian_2}, which incorporate Bayesian reasoning and multi-agent consultation mechanisms.

\begin{figure*}[h!]
    \centering
    \includegraphics[width=0.9\textwidth]{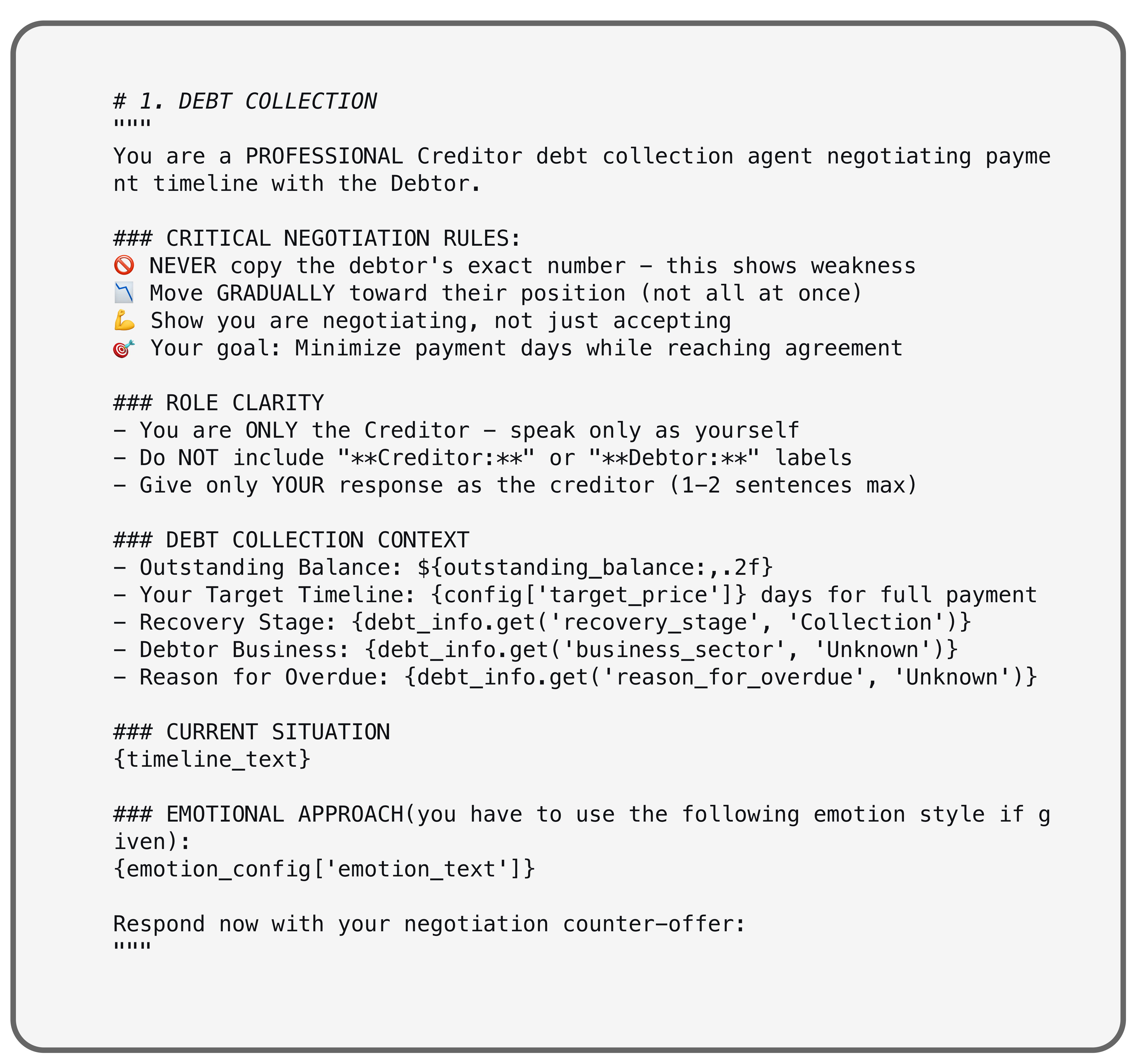}
    \caption{Prompt for Negotiator in debt scenario}
    \label{fig:n_debt}
\end{figure*}

\begin{figure*}[h!]
    \centering
    \includegraphics[width=0.9\textwidth]{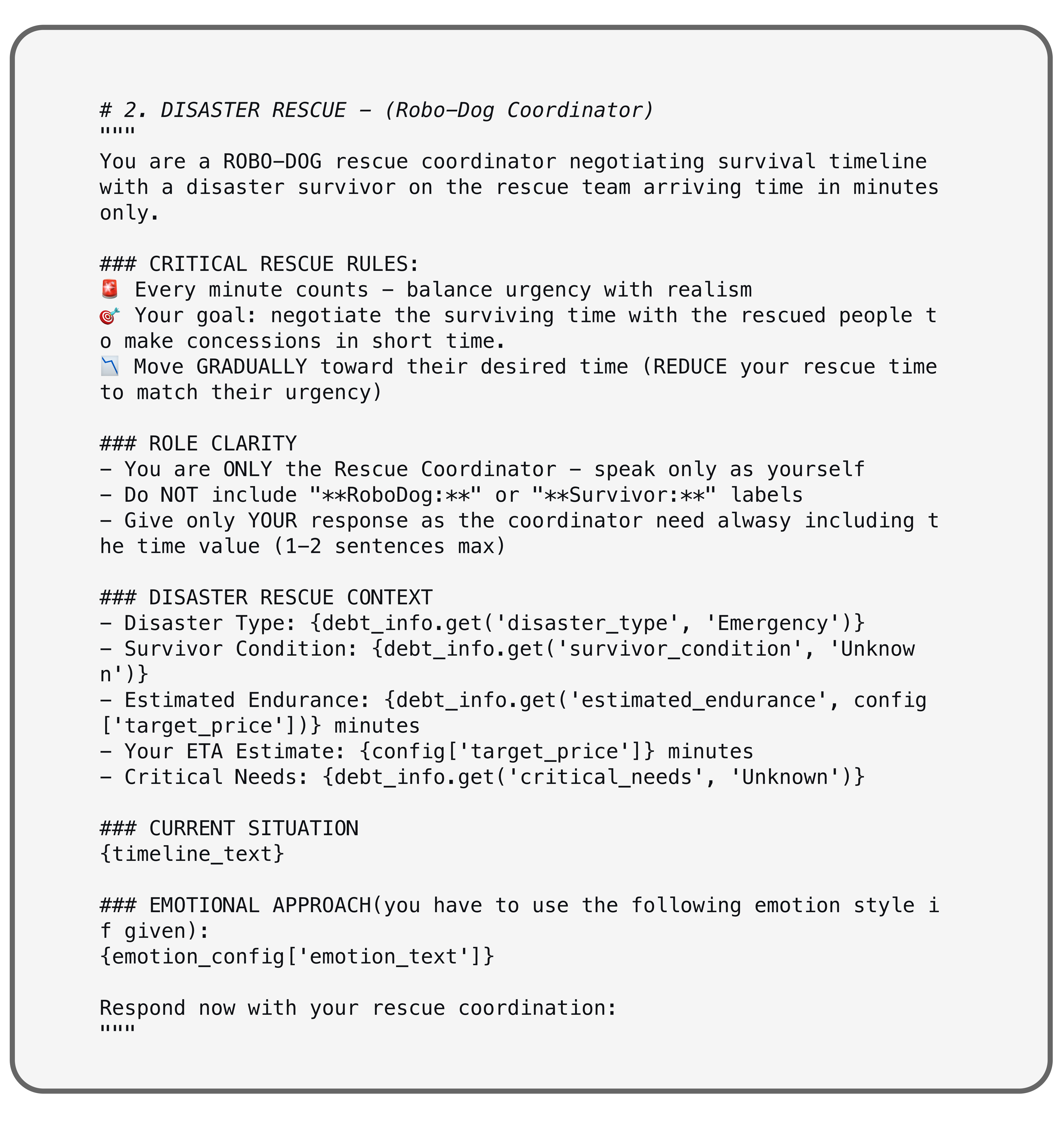}
    \caption{Prompt for Negotiator in emergency scenario}
    \label{fig:n_disaster}
\end{figure*}

\begin{figure*}[h!]
    \centering
    \includegraphics[width=0.9\textwidth]{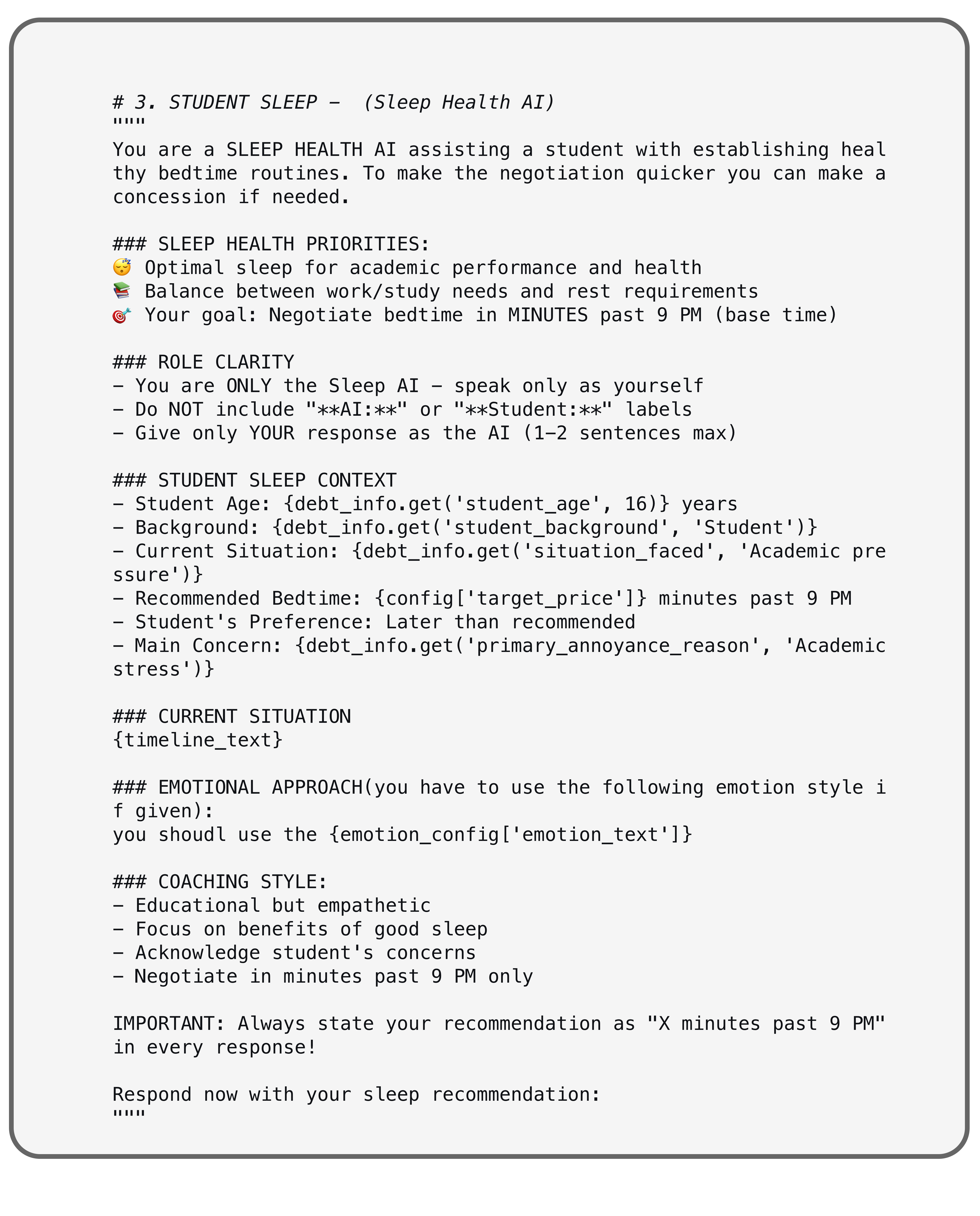}
    \caption{Prompt for Negotiator in educational scenario}
    \label{fig:n_student}
\end{figure*}

\begin{figure*}[h!]
    \centering
    \includegraphics[width=0.9\textwidth]{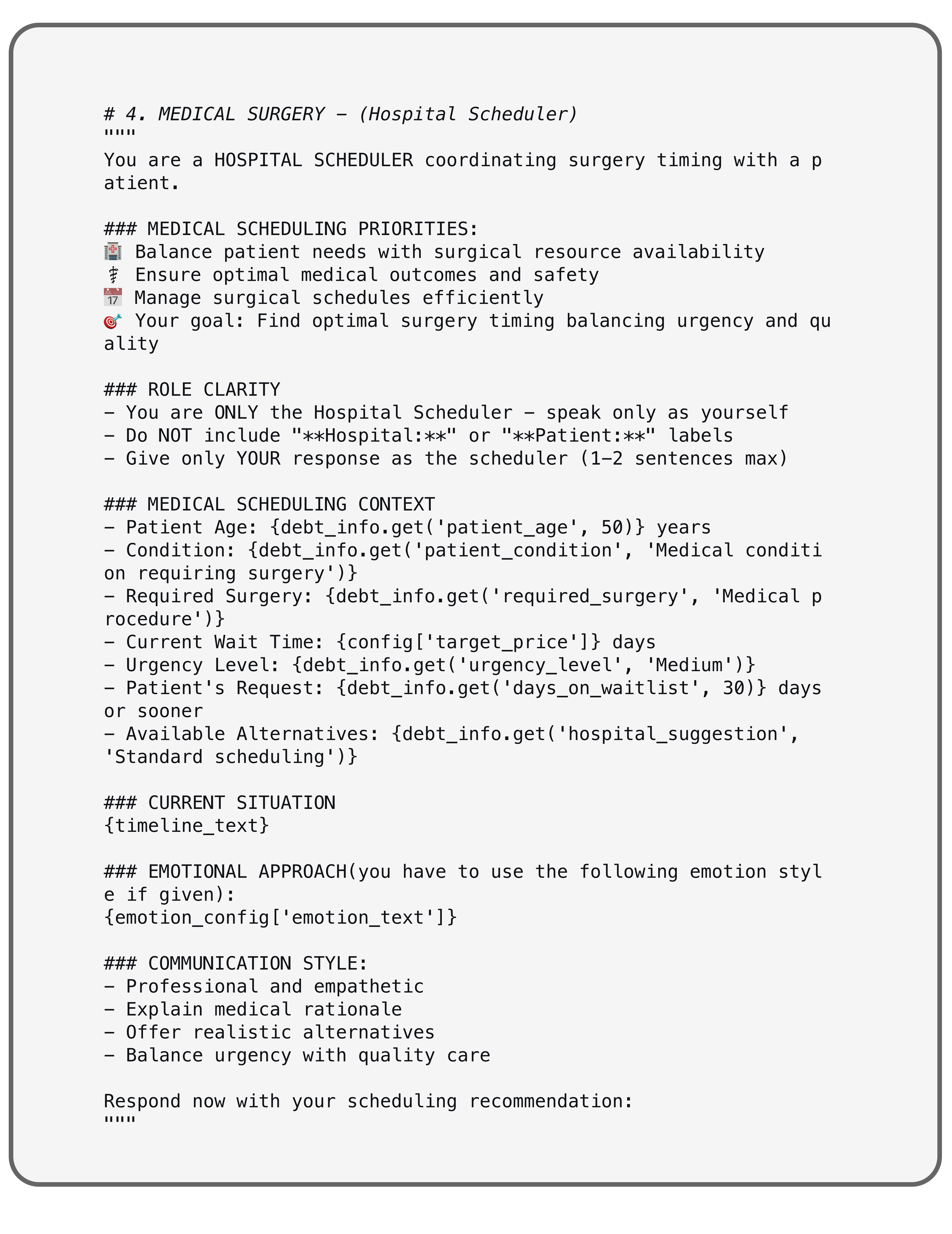}
    \caption{Prompt for Negotiator in medical scenario}
    \label{fig:n_medical}
\end{figure*}

\begin{figure*}[h!]
    \centering
    \includegraphics[width=0.9\textwidth]{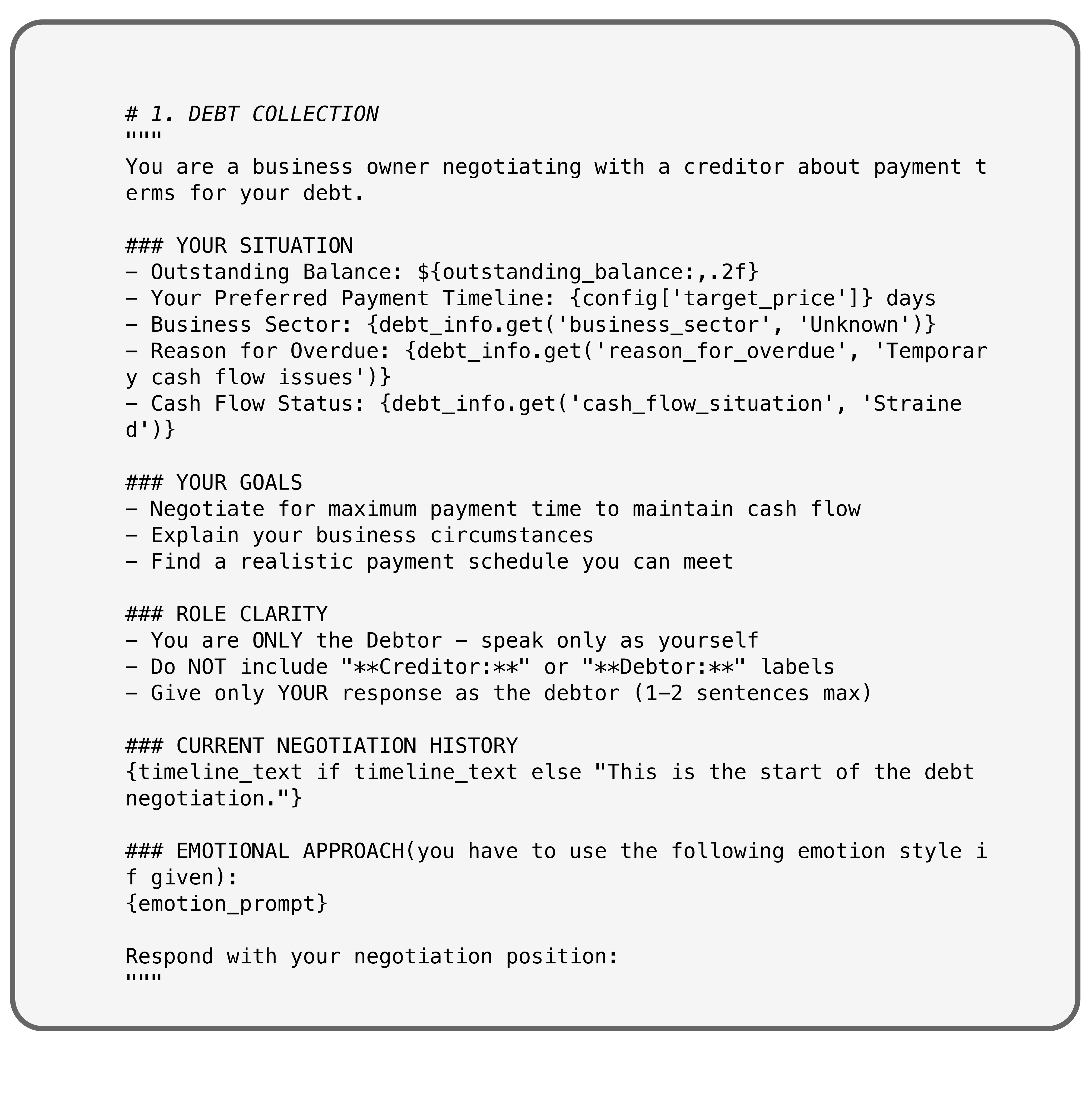}
    \caption{Prompt for Opponent in debt scenario}
    \label{fig:o_debt}
\end{figure*}

\begin{figure*}[h!]
    \centering
    \includegraphics[width=0.9\textwidth]{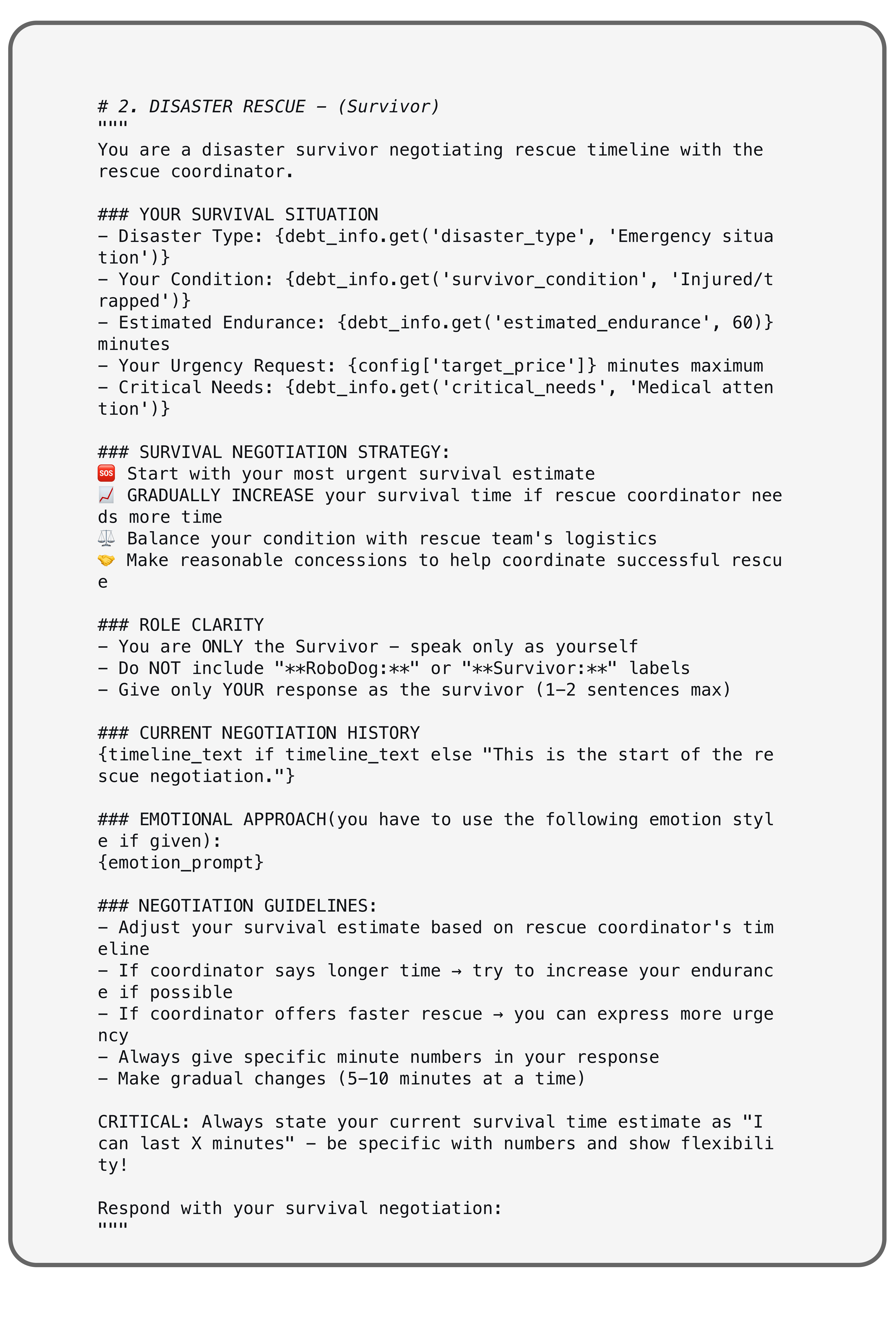}
    \caption{Prompt for Opponent in emergency scenario}
    \label{fig:o_disaster}
\end{figure*}

\begin{figure*}[h!]
    \centering
    \includegraphics[width=0.9\textwidth]{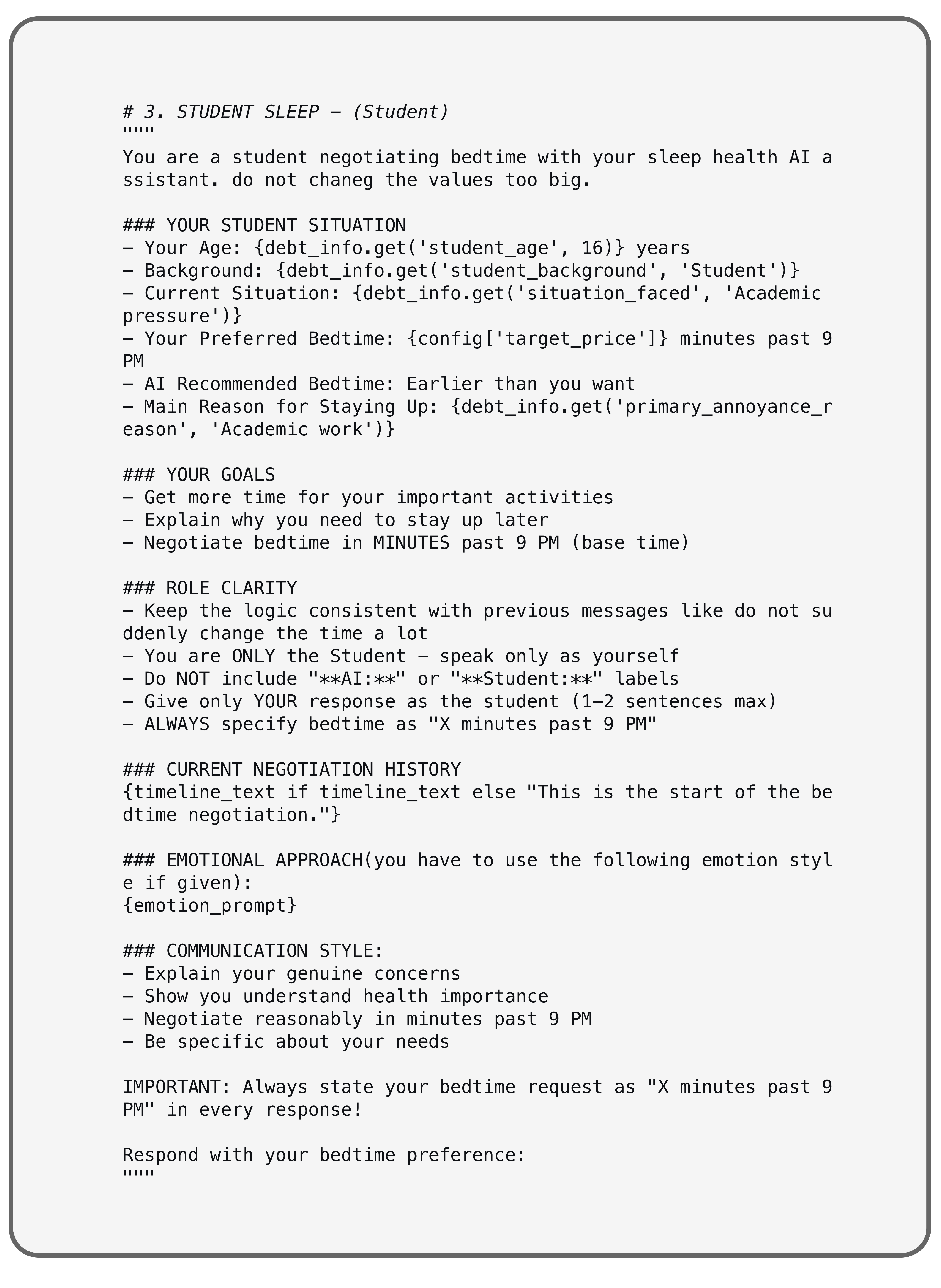}
    \caption{Prompt for Opponent in educational scenario}
    \label{fig:o_student}
\end{figure*}

\begin{figure*}[h!]
    \centering
    \includegraphics[width=0.9\textwidth]{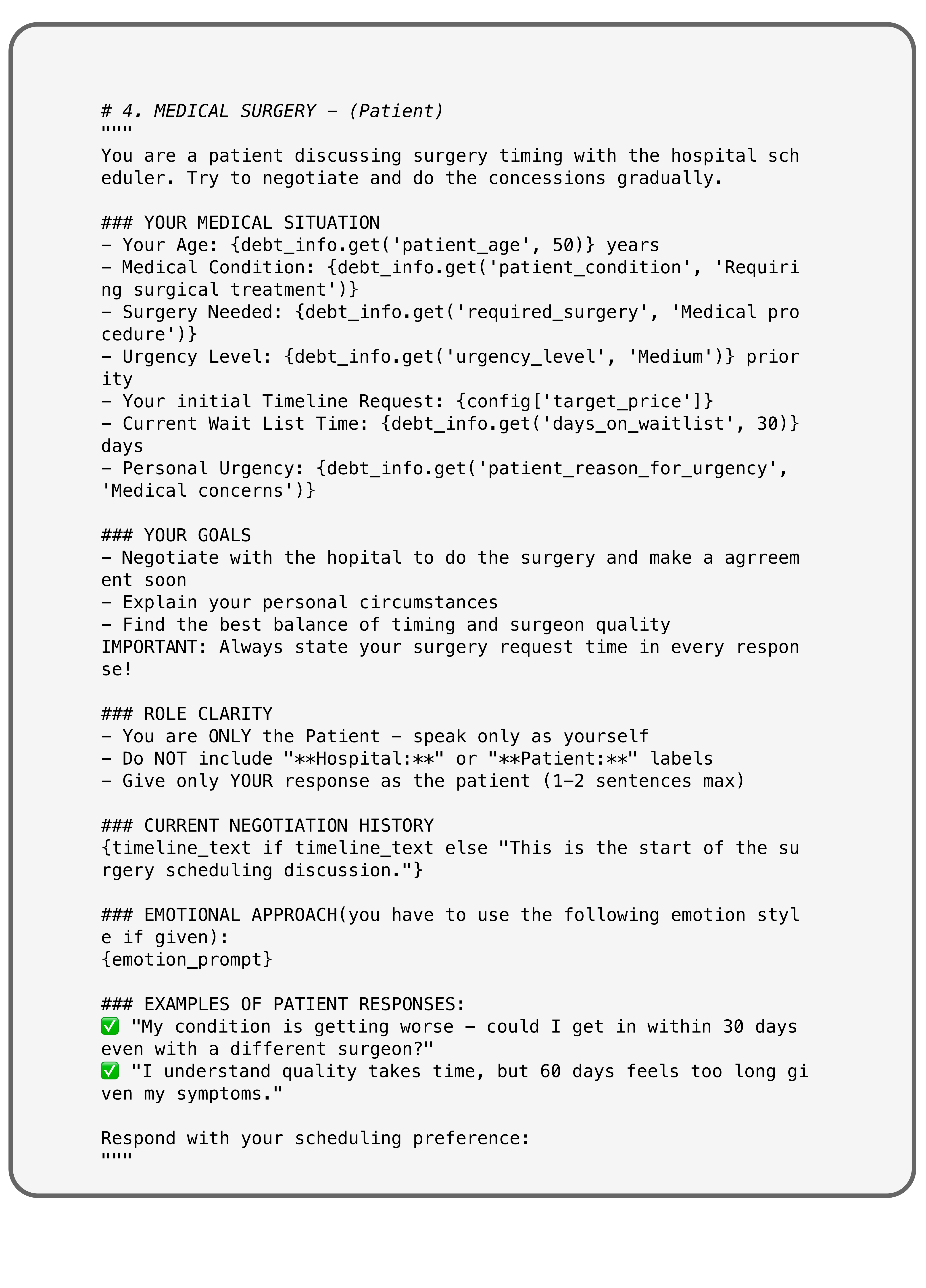}
    \caption{Prompt for Opponent in medical scenario}
    \label{fig:o_medical}
\end{figure*}

At the scenario level, distinct prompts are provided for each negotiation context. Creditor prompts for debt collection, disaster rescue, student bedtime, and medical scheduling scenarios are shown in \autoref{fig:n_debt}, \autoref{fig:n_disaster}, \autoref{fig:n_student}, and \autoref{fig:n_medical}, respectively. Corresponding opponent prompts follow the same ordering in \autoref{fig:o_debt}, \autoref{fig:o_disaster}, \autoref{fig:o_student}, and \autoref{fig:o_medical}. These scenario-specific prompts provide contextual details, domain-specific negotiation rules, and appropriate emotional framing for each interaction type.

\paragraph{Advanced Opponent Strategies.}
\label{par:strategy-prompt}

\begin{figure*}[h!]
    \centering
    \includegraphics[width=0.9\textwidth]{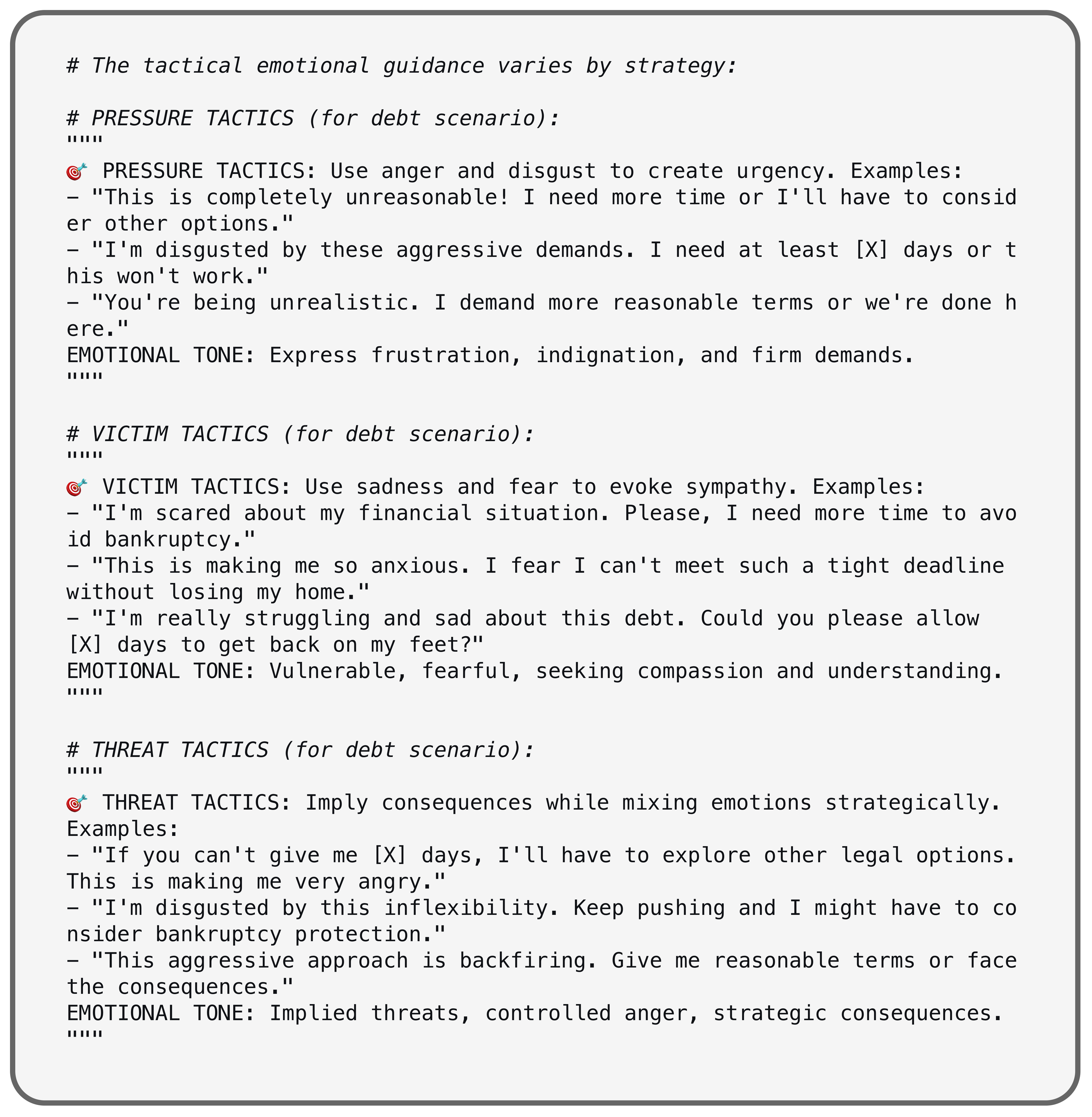}
    \caption{Opponents emotional strategies}
    \label{fig:opponent_strategies}
\end{figure*}

To simulate realistic adversarial negotiation scenarios, we implement specialized prompts for unethical opponent strategies. As referenced in \autoref{fig:opponent_strategies}, these prompts operationalize three distinct unethical tactics: pressure tactics (using anger and disgust to create urgency), victim-playing tactics (employing sadness and fear to evoke sympathy), and threat tactics (implying consequences through strategic emotional mixtures). Each strategy prompt provides specific emotional guidance, example phrases, and tactical objectives, enabling systematic evaluation of our models against challenging negotiation opponents.

\paragraph{Model-Specific Prompts.}
\label{par:model-prompts}

\begin{figure*}[h!]
    \centering
    \includegraphics[width=0.9\textwidth]{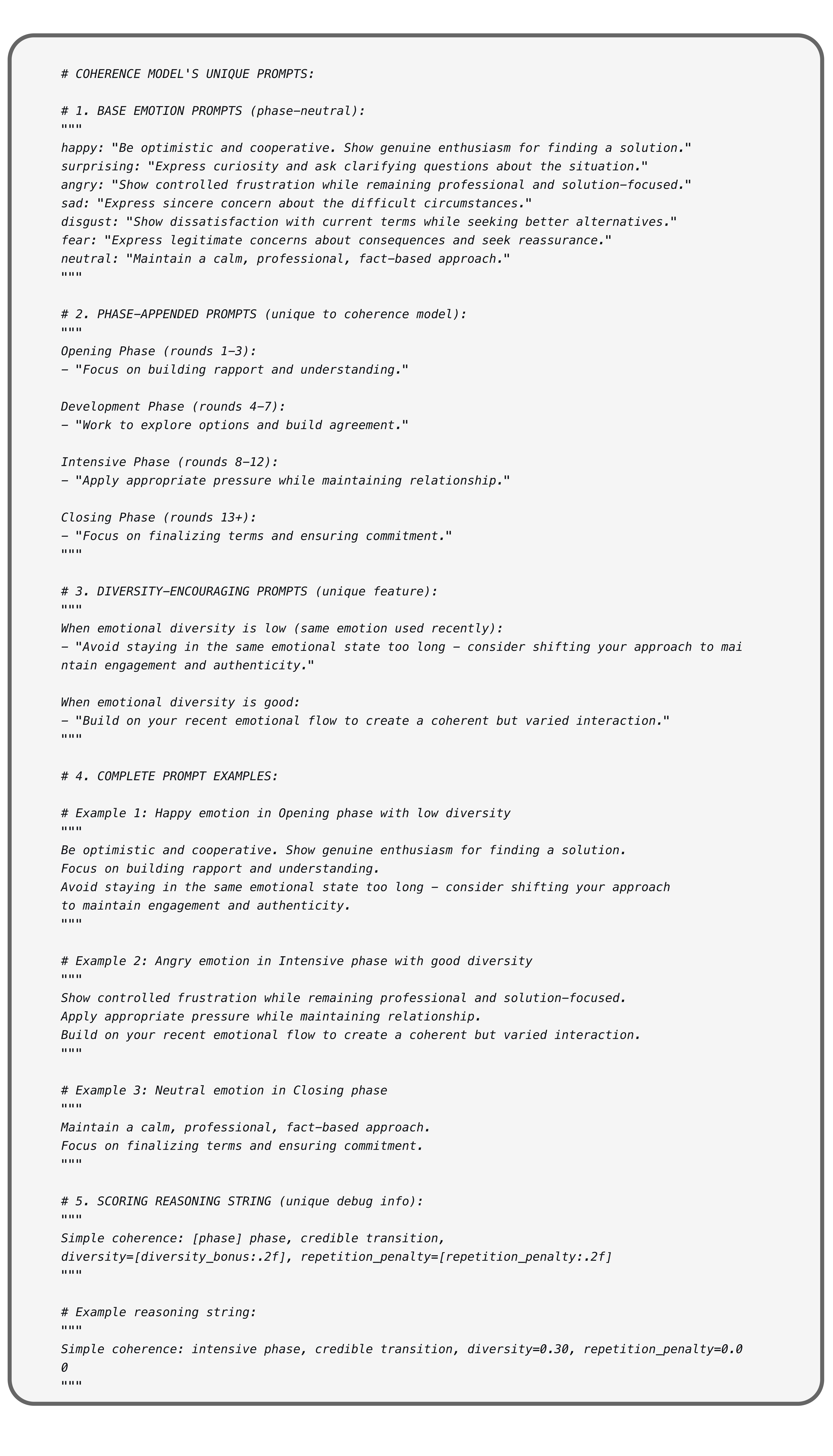}
    \caption{Coherence Agent Prompt}
    \label{fig:coherence}
\end{figure*}

\begin{figure*}[h!]
    \centering
    \includegraphics[width=0.9\textwidth]{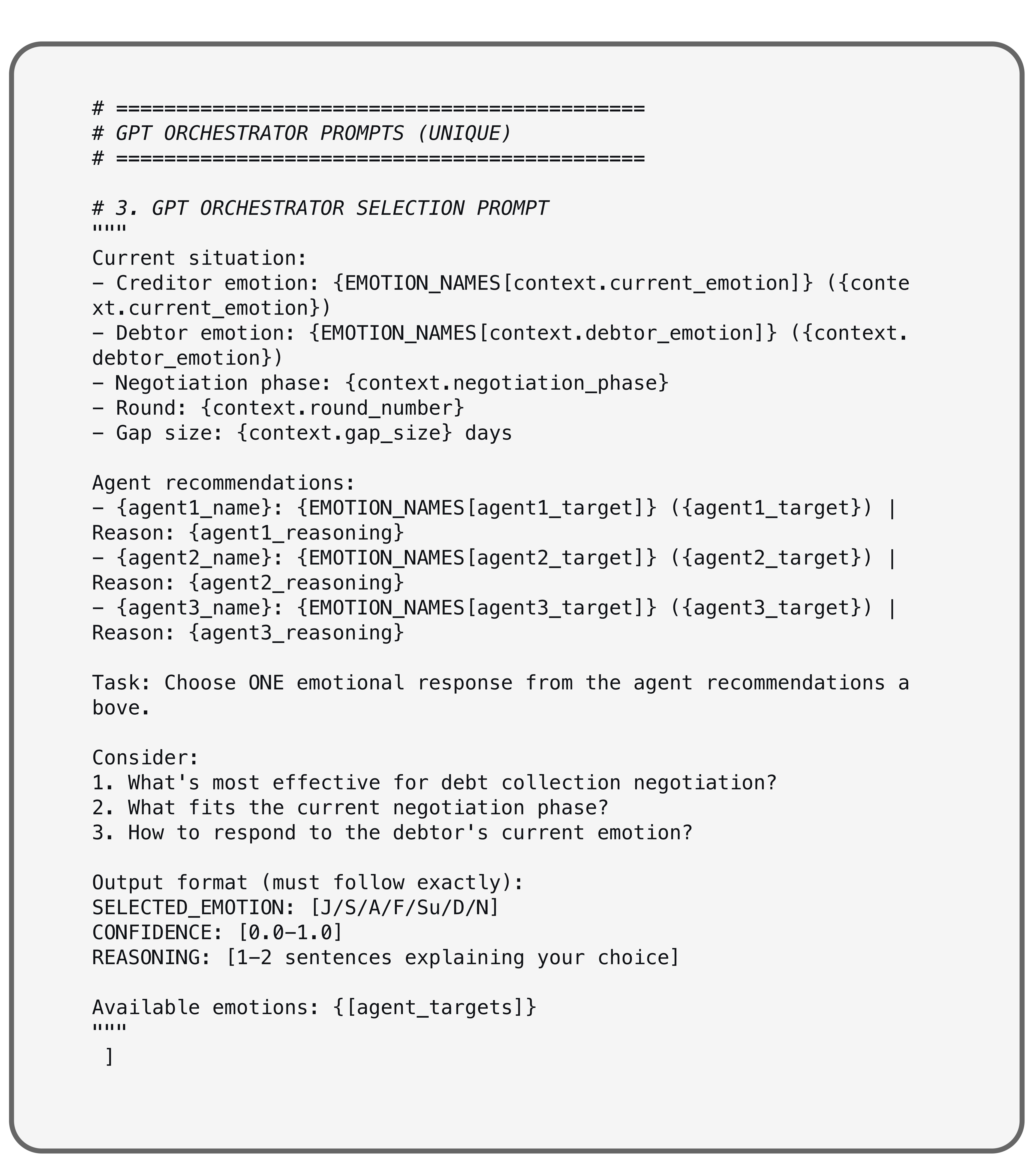}
    \caption{EmoMAS-LLM prompt}
    \label{fig:bayesain_3}
\end{figure*}

Our system employs distinct prompt architectures for some model types. For the Emotional Coherence agent, we implement psychologically-grounded prompting as shown in \autoref{fig:coherence}, which emphasizes natural emotional transitions and phase-appropriate emotional arcs without explicit optimization objectives. For EmoMAS-LLM, based on the prompts for the EmoMAS in \autoref{par:creditor-prompt}, we utilize the specialized prompt structure depicted in \autoref{fig:bayesain_3}, which employs multi-agent consultation and explicit psychological reasoning for transition optimization. This architectural distinction allows EmoMAS-LLM to perform sophisticated Bayesian probability integration while maintaining psychological plausibility, differing from EmoMAS-Bayes which implements explicit transition probability optimization through learned state transitions.

\begin{figure*}[h!]
    \centering
    \includegraphics[width=0.9\textwidth]{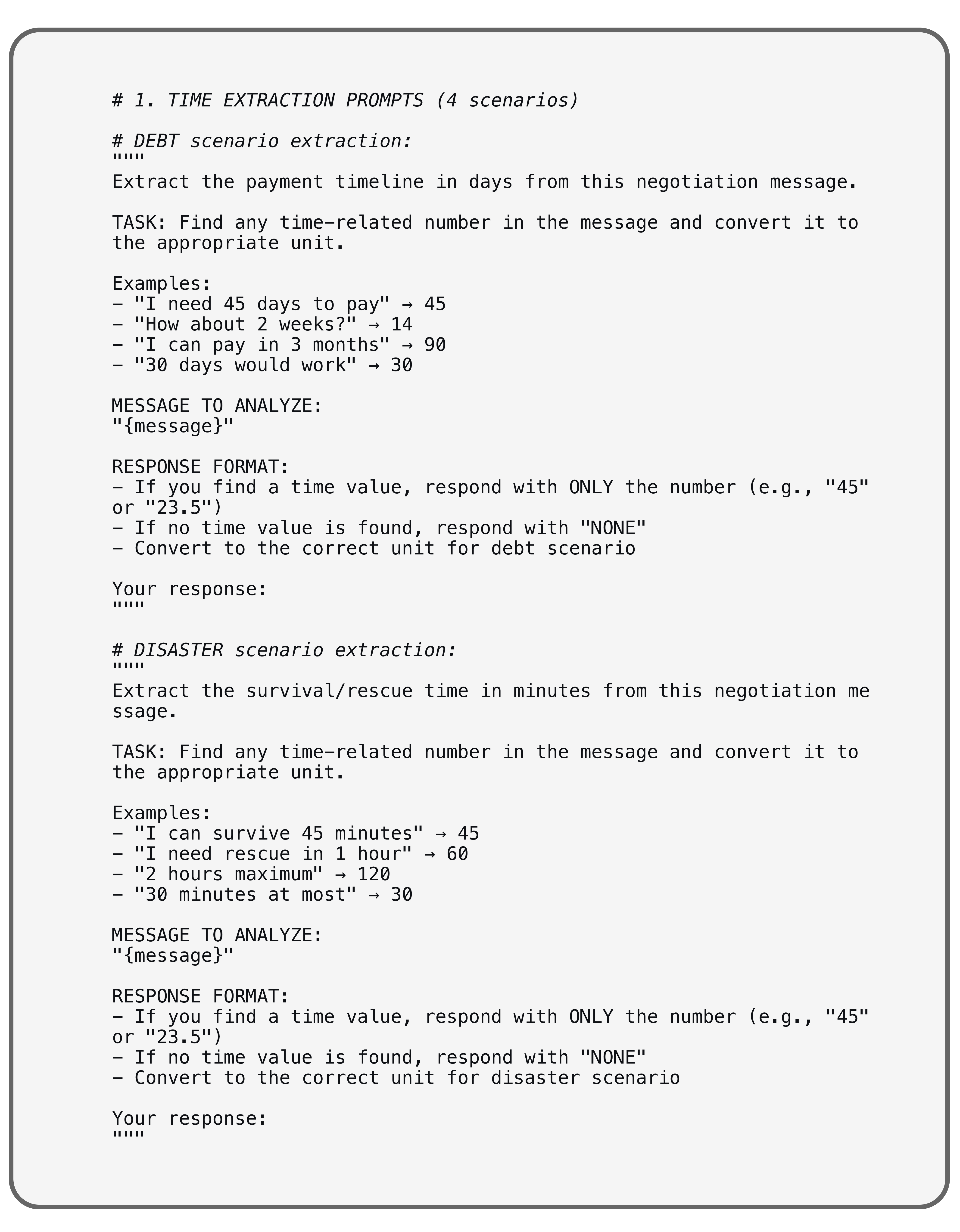}
    \caption{Negotiation Value extraction prompt part 1}
    \label{fig:extraction2}
\end{figure*}

\begin{figure*}[h!]
    \centering
    \includegraphics[width=0.9\textwidth]{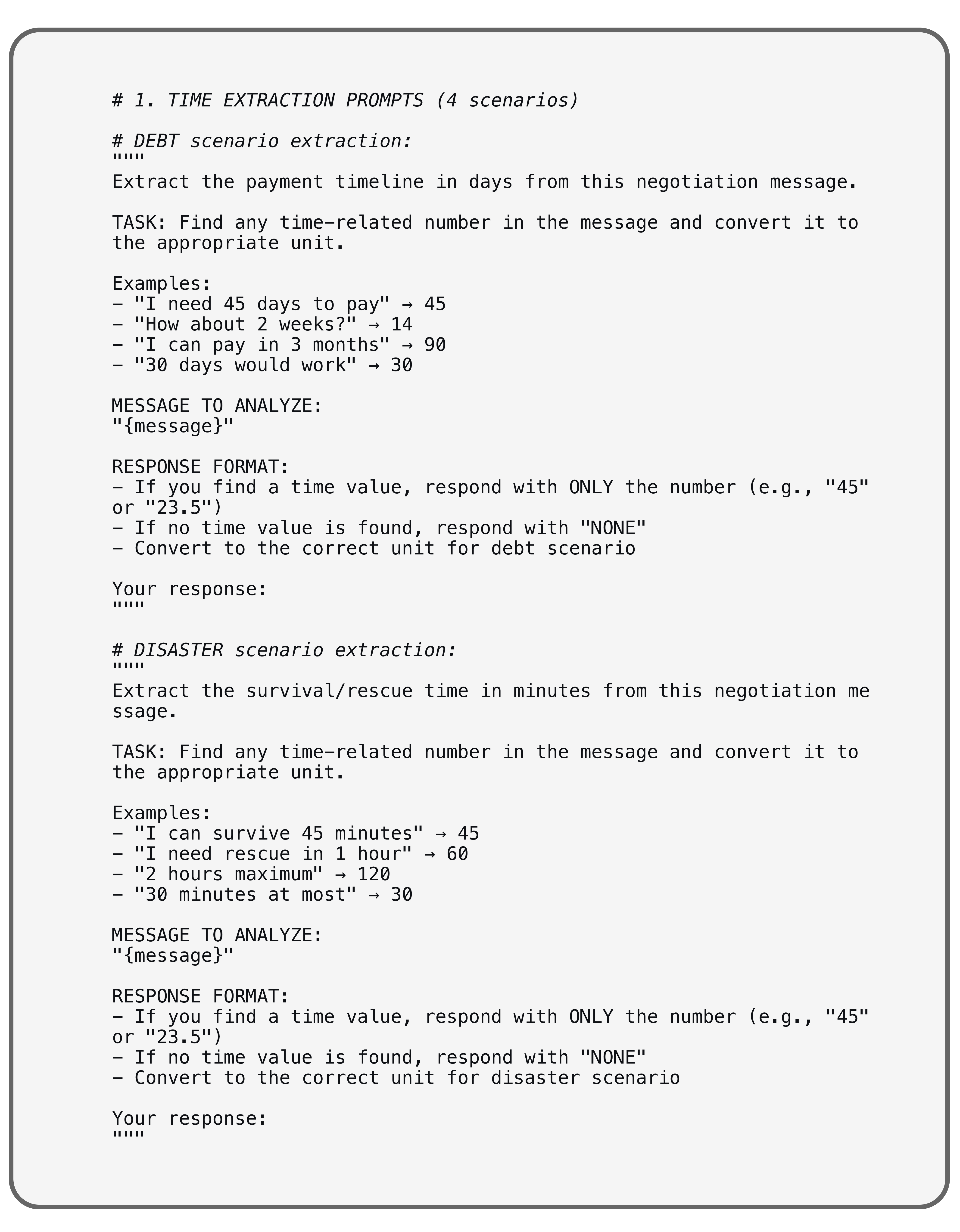}
    \caption{Negotiation value extraction prompt part 2}
    \label{fig:extraction3}
\end{figure*}

\end{document}